%% file: main.tex
\documentclass{article}

\usepackage[final]{corl_2022} % Uncomment for the camera-ready ``final'' version.
\usepackage{multirow}
\usepackage{booktabs}
\usepackage{subfigure}
\usepackage{amsmath,amsfonts,bm,amsthm,amssymb}
\usepackage{algorithm}
\usepackage{algorithmic}
\usepackage{graphics,graphicx,caption,float,color}
\usepackage{wrapfig}
\usepackage{bbm}
\usepackage{xspace}
\usepackage{colortbl}
\usepackage{appendix}
\usepackage{makecell}
 
\newlength\savewidth
\newcommand\shline{\noalign{\global\savewidth\arrayrulewidth
  \global\arrayrulewidth 1pt}\hline\noalign{\global\arrayrulewidth\savewidth}}

\usepackage{xcolor}
\definecolor{citecolor}{HTML}{0071BC}
\hypersetup{colorlinks,linkcolor={red},citecolor={citecolor}}  
\newcommand{\rowsqueeze}{\vspace{-0pt}}
\definecolor{Gray}{gray}{0.9}

\newcommand{\eg}{\emph{e.g.}}
\newcommand{\ie}{\emph{i.e.}}
\newcommand{\alias}{ECL\xspace}
\newcommand{\full}{Embodied Concept Learner\xspace}

\definecolor{MyDarkBlue}{rgb}{0,0.08,1}
\definecolor{MyDarkGreen}{rgb}{0.02,0.6,0.02}
\definecolor{MyDarkRed}{rgb}{0.8,0.02,0.02}
\definecolor{MyDarkOrange}{rgb}{0.40,0.2,0.02}
\definecolor{MyPurple}{RGB}{111,0,255}
\definecolor{MyRed}{rgb}{1.0,0.0,0.0}
\definecolor{MyGold}{rgb}{0.75,0.6,0.12}
\definecolor{MyDarkgray}{rgb}{0.66, 0.66, 0.66}
\definecolor{Gray}{gray}{0.9}
\definecolor{cssgreen}{rgb}{0.0, 0.5, 0.0}
\title{
Embodied Concept Learner: \\ Self-supervised Learning of Concepts and Mapping through Instruction Following
}

\author{%
Mingyu Ding~~~~\\
HKU\\
MIT CSAIL\\
\And
Yan Xu\\
CUHK\\
\And
Zhenfang Chen\\
MIT-IBM Watson AI Lab\\
\And
David Cox\\
MIT-IBM Watson AI Lab\\
\And
Ping Luo\\
HKU\\
\\
\And
Joshua B. Tenenbaum\\
MIT BCS, CBMM, CSAIL\\
\And
Chuang Gan\\
UMass Amherst \\ MIT-IBM Watson AI Lab\\
}
\begin{document}
\maketitle
% \citet \citep
%===============================================================================

\begin{abstract}
% Humans excel at learning concepts and skills through active interaction with the environment to achieve the natural language goal.
Humans, even at a very early age, can learn visual concepts and
% obtain experiences of 
understand
geometry and layout through active interaction with the environment, and generalize their compositions to complete tasks described by natural languages in novel scenes. To mimic such capability, we propose \full (\alias)~\footnote{Project page: \url{http://ecl.csail.mit.edu/}} in an interactive 3D environment. Specifically, a robot agent can ground visual concepts, build semantic maps and plan actions to complete tasks by learning from human demonstrations and language instructions.
% , \ie, only a pretrained object proposal network is used without dense semantic and depth supervisions from simulations. 
\alias consists of: (i) an instruction parser that translates the natural languages into executable programs; (ii) an embodied concept learner that grounds visual concepts based on language descriptions/embeddings and a pretrained object proposal network; (iii) a map constructor that estimates depth and constructs semantic maps by leveraging the learned concepts; and (iv) a program executor with deterministic policies to execute each program.
\alias has several appealing benefits thanks to its modularized design.
Firstly, it enables the robotic agent to learn semantics and depth unsupervisedly acting like babies, \eg, ground concepts through active interaction and perceive depth by disparities when moving forward.
Secondly, \alias is fully transparent and step-by-step interpretable in long-term planning.
% Thirdly, the concept learner and the map constructor benefit each other. The learned concepts provide accurate semantics for map construction; in turn, mapping corrects the wrong concepts in grounding. 
Thirdly, \alias could be beneficial for the embodied instruction following (EIF), outperforming previous works on the ALFRED benchmark when the semantic label is not provided.
Also, the learned concept can be reused for other downstream tasks, such as reasoning of object states.

\end{abstract}

% Two or three meaningful keywords should be added here

\section{Introduction}
\begin{wrapfigure}{R}{0.65\textwidth}
    \centering
    \vspace{-12pt}
    \includegraphics[width=\linewidth]{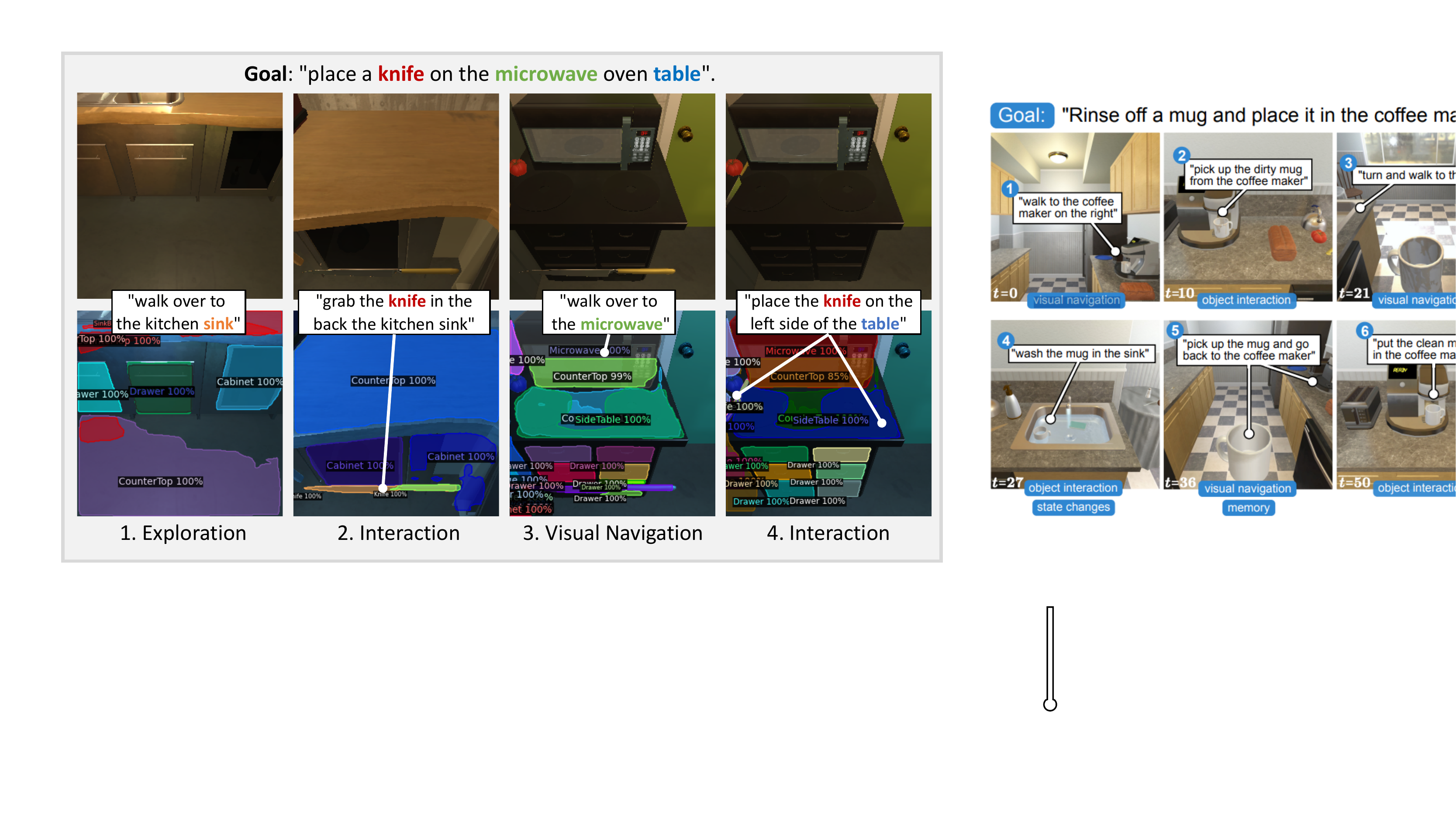}
    \vspace{-12pt}
    \caption{An example of a language goal and its corresponding four subgoals. The top and bottom rows show visual observations by the robotic agent and our grounded semantics, respectively.
    We show that we align object concepts encoded in subgoals with visual proposals to learn concepts in the embodied environment.
    % the visual trajectories on the top of the image. On the bottom, we show that how we can align object concepts encoded in the subgoals and visual proposals of the scenes to learn concepts in the embodied environments.
    }
    \label{fig:teaser}
    \vspace{-10pt}
%\end{figure}
\end{wrapfigure}

% Humans possess a versatile mechanism for exploring, navigating, and interacting in an embodied multimodal environment.
% By interacting in the environment with the help of language instructions, \eg, ``place a knife on the microwave oven table'', even a baby can understand the language concepts and their representations in the visual system, and learn compositional skills.
% To mimic such capability, the 
Embodied instruction following (EIF)~\cite{shridhar2020alfred} is a popular task in
robot learning. Given some multimodal demonstrations (natural language and egocentric vision, as shown in Fig.~\ref{fig:teaser}) in a 3D environment, a robot is required to complete novel compositional instructions in unseen scenes.
The task is challenging because it requires accurate 3D scene understanding and semantic mapping, visual navigation, and object interaction. %\zf{Should we mention the challenges of language understanding?}

%\begin{figure}[t]

Recent works for EIF can be typically divided into two streams 
and they have certain limitations.
1) End-to-end imitation learning methods~\cite{shridhar2020alfred,singh2020moca,lav,episodictransformer} directly input the visual observation of the current step and language instructions into the model, and output the action for the next step.
For example, \citet{episodictransformer} has presented the episodic transformer to predict the agent's actions with an attention mechanism and a progress monitor.
Such models work by simply memorizing training scenes and trajectories. While they achieve good performance in seen environments, they fail to generalize well in unseen scenes.
Furthermore, these black-box models often lack transparency, interpretability, and generalizability.
2) Mapping-based methods~\cite{blukis2021persistent,min2021film} leverage the map representations~\cite{walter2013learning,hemachandra2015learning,patki2019inferring,kostavelis2015semantic} by building a 3D voxel map from the predicted depths and instance segmentation masks. A semantic top-down map of the scene is then constructed and updated at each step.
These works perform explicit exploration and interactions through semantic search policies~\cite{min2021film} to achieve the natural
language goal, which is transparent and interpretable.
However, they assume that the agent has learned the depth and semantics passively from large amounts of data. 
The semantic labels and depth supervision are often labor-intensive and hard to obtain in the real world.
%Moreover, the learning strategy is in contrast to the way humans learn, which actively interacts with the environment and learns common sense in an unsupervised manner.
% {\color{red} We argue that }
We argue that such supervision signals are unnecessary since we can learn language concepts and visual disparity through interactions in the environments. For example, by achieving the goal described in Fig.~\ref{fig:teaser}, humans can learn what the concepts ``knife'' and ``table'' are and perceive that the table in frame~2 is physically closer to the agent than frame~1.
% \zf{We argue that such dense supervision signals are unnecessary since we can learn such language concepts and physical depths through interactions in the environments. For example, by achieving the goal described in Fig.~\ref{fig:teaser}, human can learn what the concepts ``kinfe'' and ``table'' are and know that frame 2 is physically closer to the agent than the frame 1.} 

This paper answers a question naturally raised from the above issues: can we make the agent behave like a baby? A baby is able to learn domain knowledge from environmental interactions and expert demonstrations without additional supervision to achieve the natural language goal.
We speculate that babies do this possibly in a way similar as:
(i) Learn skills and concepts from expert demonstrations (environment observations and language instructions),
\eg, the skill ``place'' and concepts ``knife'' and ``table'' can be grounded from the demonstration ``place a knife on the microwave oven table''.
(ii) Given a new compositional language goal like ``put a clean tomato on the dining table'' in Fig.~\ref{fig:framework}, one may process it into many subgoals, like ``pickup tomato'', ``clean tomato'', and ``put it on the table''.
(iii) Explore the scene and build a semantic map, where depth information is estimated automatically based on the disparity when moving forward or backward.
(iv) Complete each subgoal based on the learned semantic map and skills, and update the semantic map dynamically.

Motivated by the above observations, we propose \full (\alias) that mimics baby learning for embodied instruction following.
It consists of: (i) an instruction parser that parses the natural languages into executable programs; (ii) an embodied concept learner that aligns language concepts with visual instances (a pretrained object proposal network and a word embedding model are used); (iii) a map constructor based on the grounded semantic concepts and unsupervised depth estimation; and (iv) a program executor with deterministic policies to perform each subtask.
%
% These components cooperate seamlessly: the concept learner takes words from the output of the instruction parser as input;
% the concept grounding probabilities are used for Bayesian filtering in the map building and updating;
% in turn, the mapping module can correct the wrong concepts in grounding;
% a soft obstacle map is also constructed from the concept learner for the deterministic policy in the program executor.

Our contributions are three-fold. 
1) We introduce \alias, a modular framework that can ground
visual concepts, build semantic maps and plan actions to complete complex tasks by learning purely from human demonstrations and language instructions.
2) \alias achieves competitive performance without semantic labels on embodied instruction following (ALFRED)~\cite{shridhar2020alfred}, while maintaining high transparency and step-by-step interpretability.
3) We could also transfer the learned concepts to other tasks in the embodied environment, like the reasoning of object states.
% in both settings with and 

% Without semantic labels, \alias achieves state-of-the-art performance (17.2\%) by a large margin (8.7\% absolute) against the previous SOTA~\cite{episodictransformer}.
% \alias also shows competitive performance when semantic labels are provided.
% ~\zf{This sentence is confusing? It looks like that you achieve SOTA performance without using semantic labels. How about ``\alias achieves competitive performance without using any semantic labels and beats the previous SOTA method~\cite{episodictransformer} without semantic labels by a large margin (8.7\% absolute).''}
%
%
% We show the effectiveness of each component of \alias on the ALFRED~\cite{shridhar2020alfred} benchmark.
% \alias achieves competitive performance in both settings with and without semantic labels. 
% Surprisingly, it beats the previous SOTA method~\cite{episodictransformer} by a large margin (8.7\% absolute) when semantic labels are not provided.
% \alias supports learning semantic maps unsupervisedly from only demonstrations and instructions, while maintaining high transparency and interpretability like those supervised methods. 

\section{Related Work}
\noindent \textbf{Embodied Instruction Following.}
Language-guided embodied tasks have drawn much attention, including visual language navigation (VLN)~\citep{anderson2018vision, fried2018speaker, zhu2020vision, ke2019tactical, wang2019reinforced, ma2019regretful,zadaianchuk2022self}, embodied instruction following (EIF)~\cite{suglia2021embodied,lwit,abp,song2022one,hitut,shridhar2020alfred}, object goal navigation~\cite{chaplot2020object, chaplot2020learning, li2022igibson}, embodied question answering~\citep{das2018embodied, gordon2018iqa}, program sketch generation~\citep{liao2019synthesizing, trivedi2021learning}, and embodied representation learning~\citep{wang2020language, bisk2020experience, prabhudesai2020embodied}. 
Among them, EIF is one of the most challenging tasks, requiring simultaneous accurate 3D scene understanding and memory, visual navigation, and object interaction.
\cite{shridhar2020alfred,episodictransformer} present end-to-end models with an attention mechanism to process language and visual input and past trajectories, predicting the subsequent action directly.
After that, works~\cite{abp,hitut,lwit} modularly process raw language and visual inputs into structured forms by object detectors~\cite{he2017mask,ding2020learning,ding2022davit}.
The above methods lack transparency and generalizability to unseen scenes.
Recently, \cite{blukis2021persistent,min2021film} proposed mapping-based methods to convert visual semantics and estimated depth into Bird's-eye-view (BEV) semantic maps and navigate based on the spatial memory.
However, such methods require depth and semantic supervisions, hence impractical in real-world scenarios.
% We overcome the challenge by learning concepts and mapping in a self-supervised manner.
% \my{TODO}
% However, the above methods either lack transparency and generalizability to unseen scenes, or require depth and semantic supervision.
% To resolve this issue,
% Our \alias behaves more like a baby, \ie, it 

% rather than learning passively from large amounts of data as previous works do. 

% Inspired by the learning principle of humans, we follow the spirits in \cite{zhou2017unsupervised,yang2020d3vo,li2021structdepth,ji2021monoindoor} to investigate the unsupervised depth learning. We also try Bayesian Filtering for better mapping to learn the 3D layout, in contrast to previous embodied works~\cite{min2021film,blukis2021persistent} that leverage supervised depth or maps.

\noindent \textbf{Visual Grounding and Concept Learning.}
Our work is also related to visual grounding~\citep{kong2014you,plummer2015flickr30k,matuszek2012joint,karpathy2015deep,mao2016generation,zhang2018grounding,yang2021visually,chen2020cops,ding2023visual} and concept learning~\citep{Mao2019NeuroSymbolic,zfchen2021iclr,mao2021grammar,bergen2008embodied,hermann2017grounded,chen2021comphy}, which align concepts onto objects in the visual scenes. Traditional visual grounding methods~\cite{karpathy2015deep,plummer2015flickr30k} map text phrases and regional features of images into a common space for cross-modality matching.  Recently, there are some works~\citep{Mao2019NeuroSymbolic,zfchen2021iclr,ding2021dynamic} learning visual concepts through question answering in passive images or videos. Differently, we study learning both visual concepts and physical depths through language instructions in the active embodied environment, which is more similar to how humans learn in the real world. Some works study language grounding in 3D world~\cite{feng2021free,roh2022languagerefer,achlioptas2020referit3d}. 
%\my{static, no active exploration.
However, they do not involve robot agents and active exploration.
% Hermann~\etal~\citep{hermann2017grounded} 
\citet{hermann2017grounded} 
interprets language in a simple simulated 3D environment, which does not consider diverse objects and actions in challenging photorealistic environments.

\section{Method}

\begin{figure}[t]
    \centering
    \includegraphics[width=0.99\textwidth]{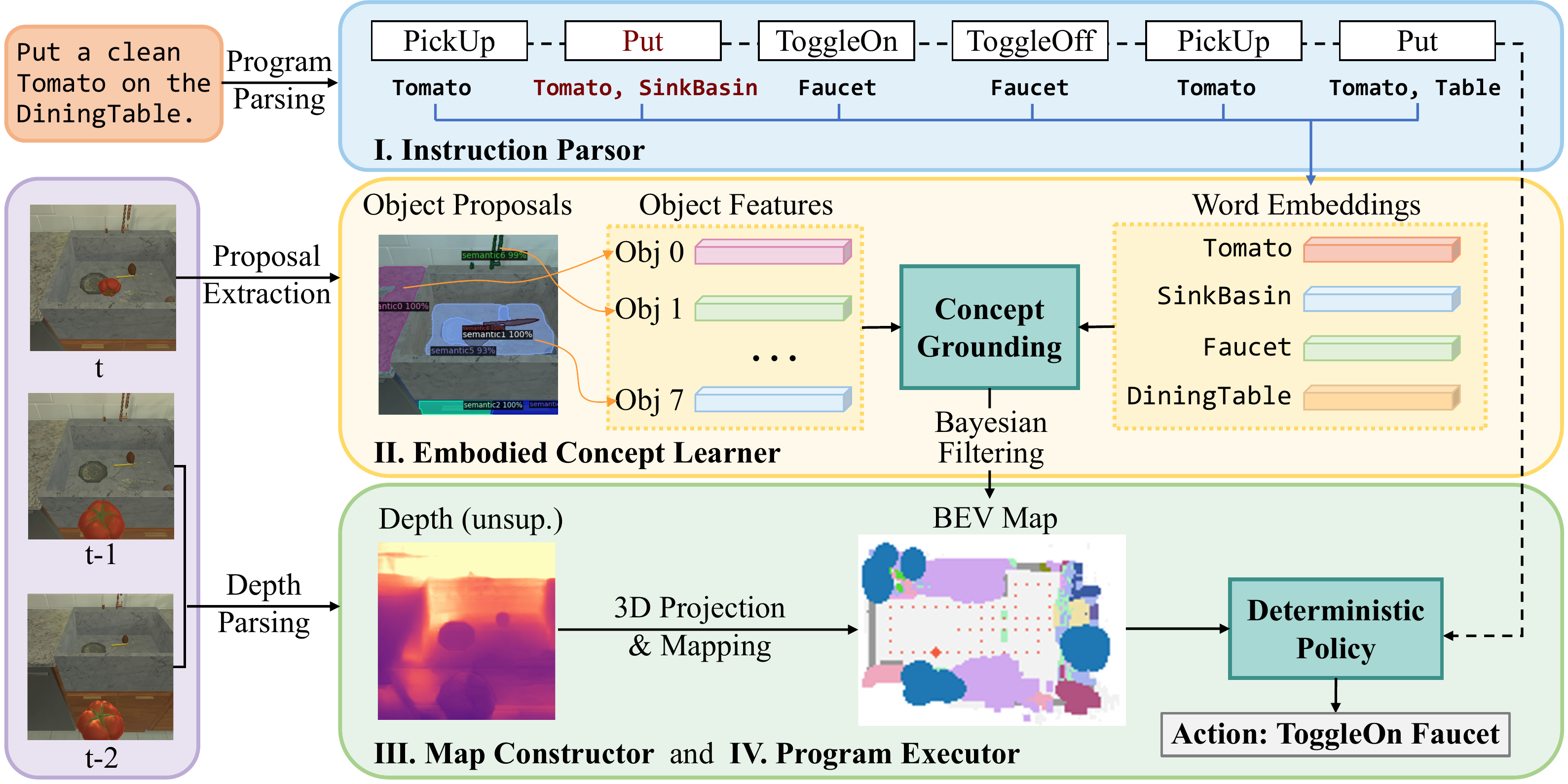}
    \caption{The framework of \alias.
    %It contains (i) an instruction parser that parses the natural languages into executable programs; (ii) an embodied concept learner that grounds visual semantics based on language descriptions; (iii) a map constructor that estimates depth and constructs semantic maps; and (iv) a program executor with deterministic policies to achieve each program.
    %\zf
    (i) Given a natural language goal, the instruction parser first parses it into a sequence of executable programs. (ii) The embodied concept learner extracts regional proposals in current frame and align them with the learned concepts. (iii) The map constructor then builds up semantic maps based on estimated depths and grounded visual concepts. (iv) 
    % With 
    Having the semantic maps and executable programs,  
    % ready, 
    the program executor predicts the agent's next action with a deterministic policy.
    }
    \vspace{-12pt}
    \label{fig:framework}
\end{figure}

In this work, we focus on the embodied instruction following task, \ie, a robotic agent is required to achieve the goal in the language instruction by exploring, navigating, and interacting with the embodied environment.
\full (\alias) includes an instruction parser, an embodied concept learner, a map constructor, and a program executor.
The modularized design ensures its transparency and step-by-step interpretability.
An overview of \alias is shown in Fig.~\ref{fig:framework}.
% In the following of this section, we describe each of its components in detail.

\subsection{Instruction Parser}
\label{sec:instruction_parser}
The instruction parser converts high-level instructions into a sequence of subtasks represented by programs.
Existing works~\cite{min2021film,blukis2021persistent,abp,hitut,episodictransformer,liao2019synthesizing,trivedi2021learning} use expert trajectories with subtasks annotations as supervision because they are easy to obtain as stated in~\cite{min2021film}.
Following this strategy,  we fine-tune a pre-trained BERT model~\citep{lewis-etal-2020-bart} learned the mapping from a high-level instruction to a sequence of subtasks (\eg, ``put a clean tomato on the diningtable'' $\rightarrow$ ``(Pickup, Tomato), (Put, SinkBasin), ...'') leveraging the subtasks sequences annotations in ALFRED~\cite{shridhar2020alfred}.
For each subtask, the instruction parser predicts the arguments, which are the same as in \cite{min2021film}: (i) ``obj'' for the object to be picked up, (ii) ``recep'' for the receptacle where ``obj'' should be ultimately placed, (iii) ``sliced'' for whether \textit{``obj''} should be sliced, and (iv) ``parent'' for tasks with intermediate movable receptacles (\eg, ``cup'' in ``Put a knife in a cup on the table''). 
% An object in ALFRED is always an instance of either \textit{``obj''} or \textit{``recep''}; \textit{``parent''} objects are a subset of \textit{``recep''} objects that are movable. We train a separate BERT model for each argument predictor.
%
After we get the subtask programs, we extract the language embeddings $e\in \mathbb{R}^{768}$ of the object words in all subprograms through a pretrained Bert model (bert-base-uncased)~\cite{meelfy} for the follow-up concept learner module.

\subsection{Embodied Concept Learner}
\label{sec:concept_learner}
% The embodied concept learner grounds visual concepts based on the language description of subgoals in an unsupervised manner.
% Let us take a look at how humans learn concepts.
Humans, even at a very early age, naturally perceive and parse the scene as objects for further understanding, \ie, grouping pixels to regions without knowing their semantics~\cite{carbon2014understanding,regan2000human}.
They then learn the object concepts from active interactions or expert demonstrations.
Similarly, the embodied concept learner leverages an object proposal network~\cite{he2017mask} without category labels
% Mask R-CNN~\cite{he2017mask} without category labels as the object proposal network. We then 
and grounds the object semantics from subgoal programs.
There are two cases to be considered:
1) If a subgoal completes, the object and its corresponding receptacle objects must be displayed in the current visual frame, and most likely in adjacent frames. In this way, the concept of these objects can be grounded.
% 1-3
For example, ``go to \underline{microwave}'', ``put the \underline{mug} on the \underline{coffeemachine}'', and ``put a \underline{mug} with a \underline{pen} in it on the \underline{shelf}'' involve 1, 2, and 3 objects, respectively.
We sample visual data from four frames before completing the subtask and two frames after it to learn the visual concept based on the corresponding action descriptions.
% These visual sequences correspond to the word embeddings involved in the subgoal.
2) If the robot agent acts ``Pickup an object'', the object appears in visual observation until the robot drops it.
% We use those visual frames and action descriptions to learn the visual concept of this object. eventually mixed
The two types of interaction data are merged and shuffled and used as input to our embodied concept learner. 
% \textcolor{red}{Note that the assumptions based on the artifact of the AI2THOR environment could be applied to real environments (when a robot picks up an object, the object appears in visual observation).}

Concretely, let $\{o_1, o_2, ..., o_k\}$ denotes $k$ objects detected in an visual input, and $\{f_1, f_2, ...f_k\}$ is their corresponding feature representations from the last layer of the object proposal network ($f \in \mathbb{R}^{1024}$ ).
Let $\{e_1, e_2, ..., e_l\}$ represents $l$ word embeddings in a subgoal (program representation, $e \in \mathbb{R}^{768}$, stated in Sec.~\ref{sec:instruction_parser}).
We first project the visual representation $f$ into the semantic space $f'\in \mathbb{R}^{768}$ where the word embeddings reside by a two-layer perceptron (MLPs). The MLPs have dimensions of $1024 \rightarrow 1024 \rightarrow 768$ with Layer Normalization~\cite{ba2016layer} and GELU activation~\cite{hendrycks2016gaussian} between the two layers.
%
% lateral chamfer distance.
We then leverage the Hungarian maximum matching algorithm~\cite{kuhn1955hungarian} for the $k$-$l$ matching, and a $\mathrm{min}(k, l)$ object visual representations can be matched with their word embeddings.
Given an assignment matrix $x\in \mathbb{R}^{k \times l}$, the task could be formulated as a minimum cost assignment problem mathematically as follows:
\begin{align}
\min_x\sum_{i=1}^k\sum_{j=1}^l d(f'_i, e_j) x_{ij}~~~\text{s.t.}~~~\sum_{i=1}^k x_{ij}=1, \sum_{j=1}^l x_{ij}\in \{0,1\}, x_{ij} \in \{0,1\},
\end{align}
where $d(\cdot)$ denotes the mean square error (MSE) and we assume $l < k$ here, vice versa.
%
% In practice, we have:
% \begin{align}
% a_1 = \mathrm{argmin}&\left( \{d(f'_i, e_1)\}_{i\in [1,k]} \right), x[a_1, 1] = 1 \nonumber \\
% a_2 = \mathrm{argmin}&\left( \{d(f'_i, e_2)\}_{i\in [1,k]\setminus\{a_1\}}\right), x[a_2, 2] = 1\\
%  & ... \nonumber
% \end{align}
% where $a_j$ is the id of the object proposal that most greedily matches the j-th word vector, and we compute the loss after $x$ is determined.
In this way, we solve a min-min optimization problem by Hungarian matching, where the first minimization is used to find the best match among the two sets of features (Hungarian matching); and the second minimization is to optimize a smaller L2 loss on the matching for learning better projected representation $f'$. Both the mapping function (MLPs) and the matching are learned at the same time. Thus the MLP and the matching matrix $x$ is jointly learned from Hungarian matching.% in a differentiable manner.
% We compute the loss after $x$ is determined to learn the semantic projection model.

During inference, we project each object proposal representation into the semantic space and perform nearest neighbor search (NNS) to assign a category label for it.
We also calculate a soft class probability $p_i$ for the i-th object by $\mathrm{softmax}\left(\{ 0.1/d_{ij} \}_j\right)$, where $d_{ij}$ is the retrieval distance between the i-th object feature and the j-th word embedding.
%distribution
The semantic probability $\mathbf{p}$ will be used for 1) Bayesian filtering in mapping and 2) statistics of the most likely location of each type of object as a navigation policy.

\subsection{Map Constructor}
% \noindent\textbf{Unsupervised depth perception learning.} 
% One important way 
Human beings understand the semantics and layouts of space, \eg, a room, mainly by first moving around, then perceiving the depth (geometry), and finally building up a semantic virtual map in our mind~\cite{maguire1998knowing}. To mimic this process, we propose a semantic map construction module leveraging the unsupervised depth learning technique~\cite{godard2019digging,zhou2017unsupervised} and probabilistic mapping inspired by Bayesian filtering. 
Concretely, we first train a monocular depth estimation network unsupervisedly, leveraging the photometric consistency~\cite{godard2019digging} among adjacent RGB observations
captured by a roaming agent. 
We use the unsupervised depth estimation 
% from this network 
for map construction. 
To build up the map, we represent the scene as voxels. Each voxel 
% at the location $\mathbf{x}_v$ 
maintains a semantic probability vector $\mathbf{p}_v$ (obtained from Sec.~\ref{sec:concept_learner}) and a scalar variable $\sigma_v$ that represents the measurement uncertainty of this voxel. 
As the new depth observation come in, we first project it to 3D space as a 3D point cloud and then transform it into the map space according to the agent ego-motion.
The transformed point cloud is voxelized for the follow-up map fusion. 
We denote the newly observed point clouds (after voxelization) as $S=\{(\mathbf{p}_s, \sigma_s )\}_{s=1}^{|S|}$ and the current voxel map as $M=\{(\mathbf{p}_m, \sigma_m)\}_{m=1}^{|M|}$. 
The newly observed voxels are fused to update the previous map as:
\begin{equation}
   \mathbf{p}_m \leftarrow \frac{\sigma_s^2}{\sigma_s^2+\sigma_m^2} \mathbf{p}_m + \frac{\sigma_m^2}{\sigma_s^2+\sigma_m^2} \mathbf{p}_s, ~\sigma_m \leftarrow (\sigma_s^{-2}+\sigma_m^{-2})^{-\frac{1}{2}}.
\end{equation}
Here, we assume $\mathbf{p}_s$ and $\mathbf{p}_m$ are 
% the coordinates of 
the semantic log probability vectors (obtained from Sec.~\ref{sec:concept_learner}) belonging to  
a pair of corresponding voxels in the new frame and the current map respectively. $\sigma_s$ and $\sigma_m$ are the estimated variances of these two voxels. Initially, the variance $\sigma_s$ of the observed voxel is predicted by a CNN. This CNN is trained with the depth estimation network in an unsupervised manner by assuming a Gaussian noise model following~\cite{kendall2017uncertainties}. 
The uncertainty-aware mapping makes it possible to correct previous mapping errors as the exploration goes on. Our probabilistic mapping is proven to be essential especially when the depth measurements are erroneous.

\subsection{Program Executor}

After concept learning and mapping, we take the average semantic probability map from demonstrations as our navigation policy. It indicates the location where each type of object most likely exists.
Although the previous work FILM~\cite{min2021film} trains a semantic policy model to predict the possible location of an object given a part of the semantic layout, the model is likely to be over-fitting. In contrast, our semantic policy is the averaged semantic map based on statistics without training, producing stable results.
% The program executor executes each program based on the current semantic map, the semantic policy (averaged semantic map), and the deterministic policy.
As shown in Fig.~\ref{fig:framework}, given the predicted subprogram, the current semantic map, and a search goal sampled from the semantic policy (averaged semantic map), the deterministic policy outputs a navigation or interaction action. 

The deterministic policy is defined as follows. If the object needed in the current subtask is observed in the current semantic map, the location of the object is selected as the goal; otherwise, we sample the location based on the distribution of the corresponding object class in our averaged semantic map as the goal.
The robot agent then navigates towards the goal via the Fast Marching Method \citep{Sethian1591} and performs the required interaction actions.

\input{tables/main_tbl}

\begin{figure}[t]
\begin{minipage}[t]{0.485\linewidth}
\begin{figure}[H]
    \centering
    \includegraphics[width=\linewidth]{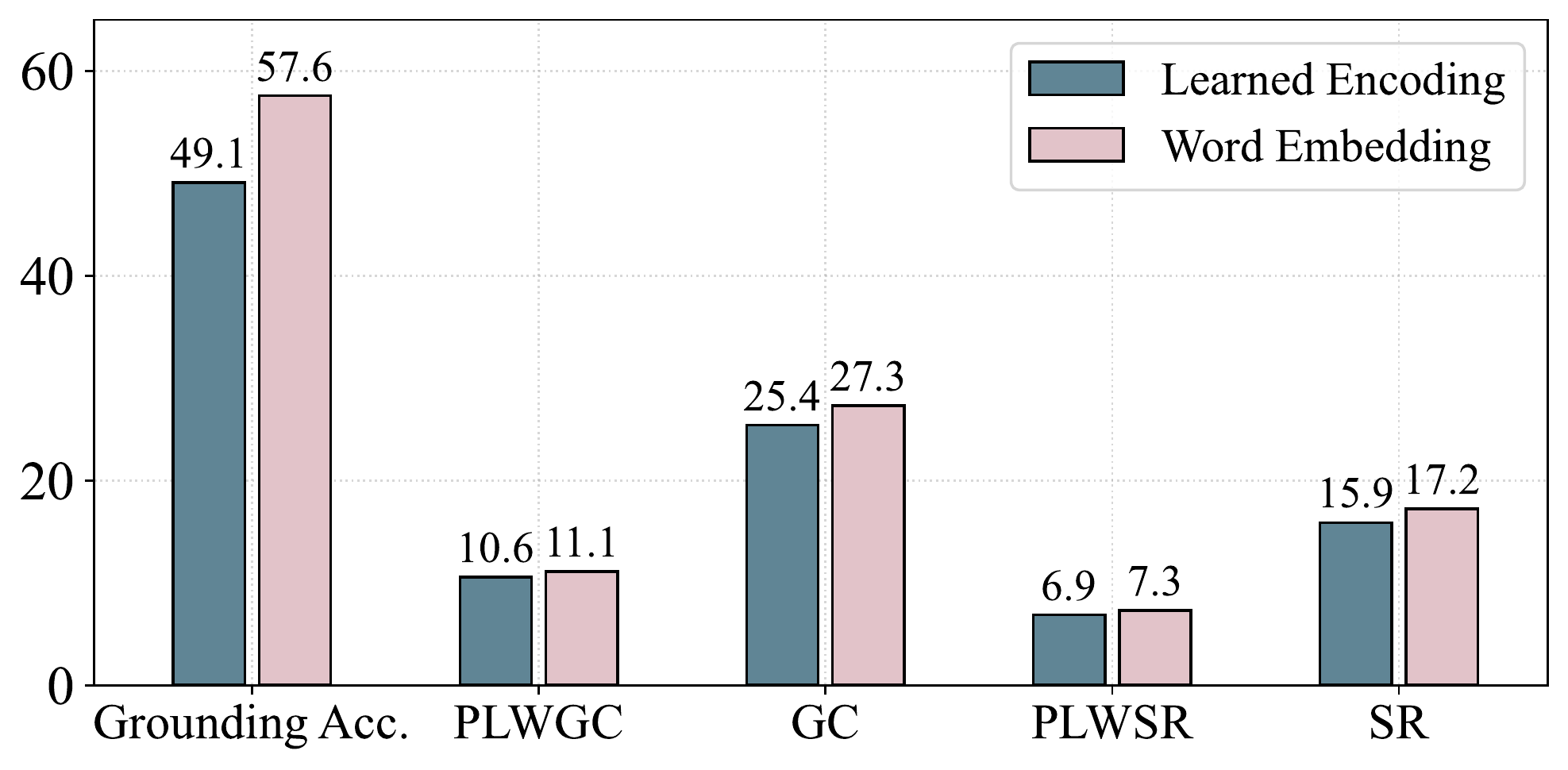}
    \vspace{-16pt}
    \caption{Results with different language representations in concept learning on \texttt{test\_unseen}.}
    \label{fig:ablation_concept}
\end{figure}
\end{minipage}
\hfill
\begin{minipage}[t]{0.485\linewidth}
\begin{figure}[H]
    \centering
    \includegraphics[width=\linewidth]{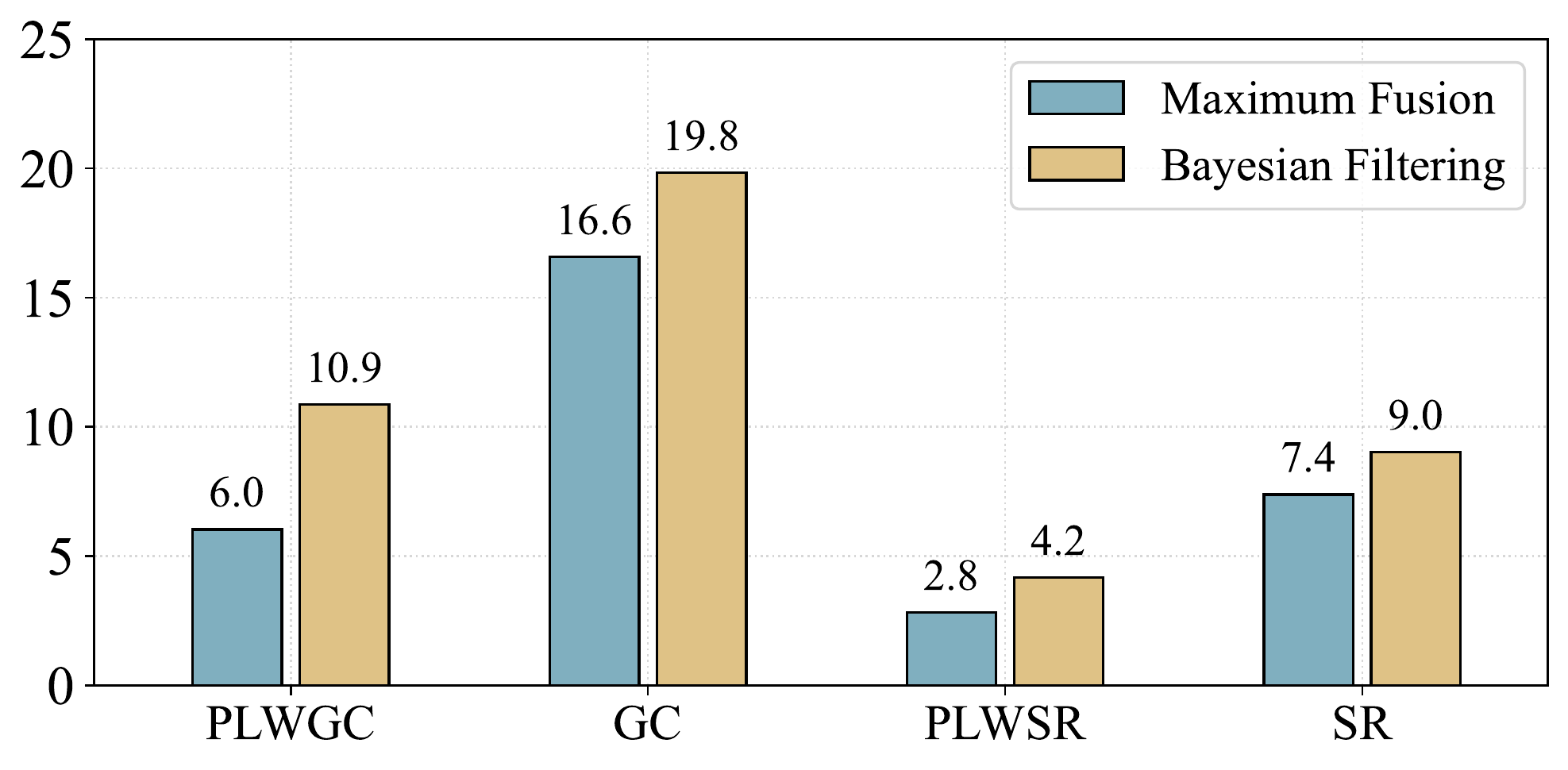}
    \vspace{-16pt}
    \caption{Evaluation with different 
    % map fusion strategies for 
    semantic mapping techniques on \texttt{test\_unseen}.}
    \label{fig:ablation_mapping}
\end{figure}
\end{minipage}
\vspace{-16pt}
\end{figure}

% \begin{wrapfigure}{R}{0.4\textwidth}
%     \centering
%     % \vspace{-12pt}
%     \vspace{-5ex}
%     \includegraphics[width=\linewidth,trim=0cm 0cm 0cm 0cm, clip]{figures/ground_acc/concept_small_h.pdf}
%     \vspace{-18pt}
%     \caption{
%     Object grounding accuracy with our concept learning on small objects. 
%     % Per-class concept learning accuracy. 
%     % (ordered). 
%     Results for challenging small objects are shown here. Please refer to the appendix for complete analysis. 
%     % More results are included in the Appendix. 
%     % and receptacle objects are shown in Appendix. 
%     % We show 23 small objects 
%     % % (could be picked up) 
%     % and the overall accuracy here, more results and the receptacle objects are shown in Appendix.
%     }
%   \label{fig:concept_small}
%     % \vspace{-20pt}
%     \vspace{-9ex}
% \end{wrapfigure}

% \begin{figure}[t]
%     \centering
% \includegraphics[width=1\linewidth,trim=0cm 10cm 9cm 0cm, clip]{figures/ground_acc/concept_learning_acc.pdf}
%     \vspace{-12pt}
%     \caption{Per-class concept learning accuracy (ordered). We show 23 small objects (could be picked up) and the overall accuracy here, more results and the receptacle objects are shown in Appendix.}
%     \vspace{-12pt}
%     \label{fig:concept_small}
% \end{figure}

\section{Experiments}
We show the effectiveness of each component of \alias on the ALFRED~\cite{shridhar2020alfred} benchmark.
For the EIF task, we report Success Rate (SR), goal-condition success (GC), path length weighted SR (PLWSR), and path length weighted GC (PLWGC) as the evaluation metrics on both seen and unseen environments.
% \noindent \textbf{Metrics.}
SR is a binary indicator of whether all subtasks were completed. GC denotes the ratio of goal conditions completed at the end of an episode. 
Both SR and GC can be weighted by (path length of the expert trajectory)/(path length taken by the agent), which are called PLWSR and PLWGC.
We also report the (grounding) accuracy for the concept learning and downstream reasoning tasks.
More details of the benchmark and the training settings for each component can be found in Appendix.

\subsection{Embodied Instruction Following on ALFRED}
\begin{figure}[t]
    \centering
\includegraphics[width=0.98\linewidth]{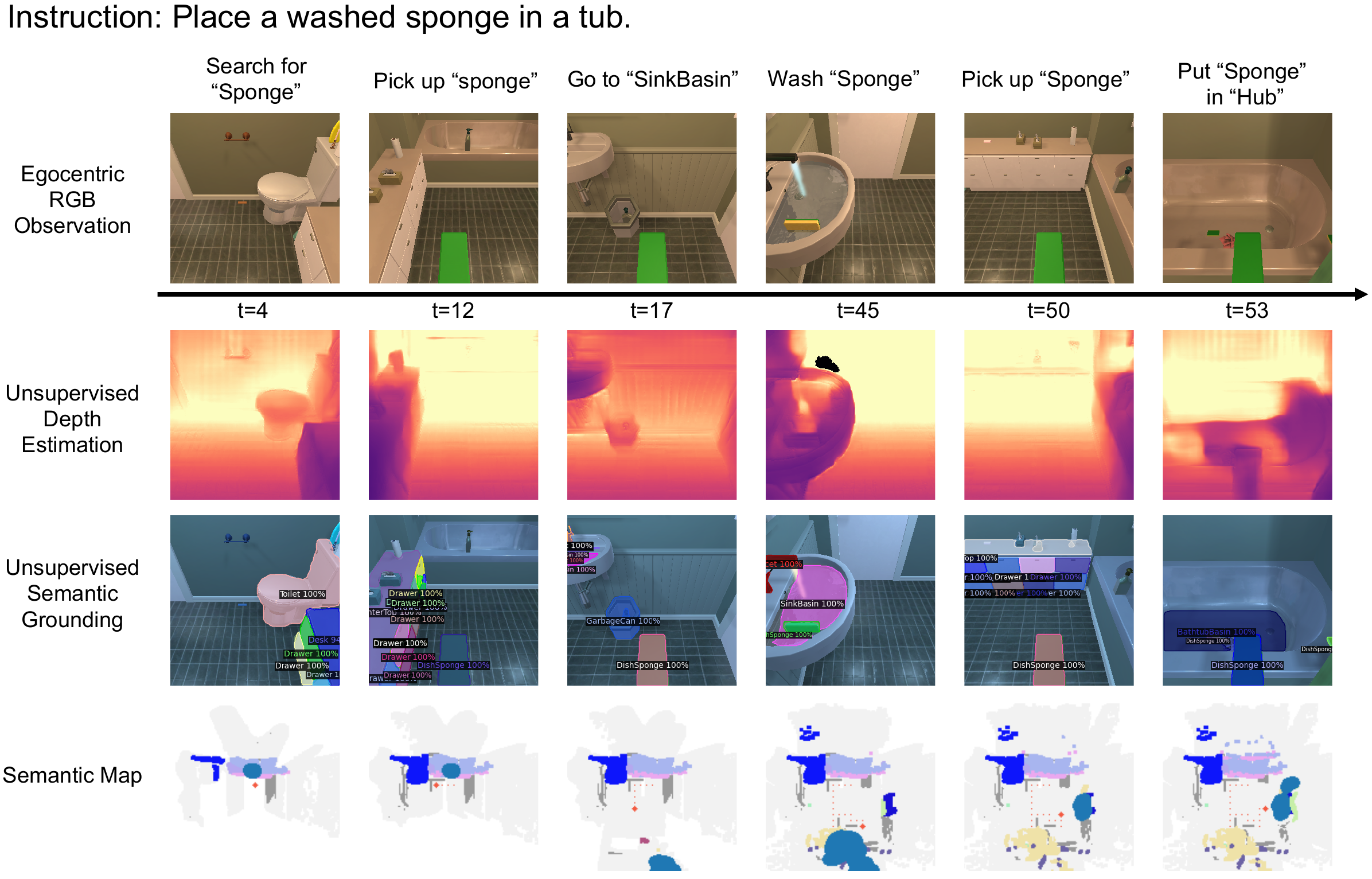}
    \vspace{-6pt}
    \caption{Visualization of intermediate estimates by \alias when an agent tries to accomplish an instruction. Based on the RGB observations, our system estimates the depths and semantic masks. The BEV semantic map is gradually established with these estimates as the exploration goes on.  
    The goals (sub goal/final goal) are represented by big blue dots in the semantic map, while the agent trajectories are plotted as small red dots.
    }
    \vspace{-18pt}
    \label{fig:instruction_vis}
\end{figure}

The results on ALFRED are shown in Tab.~\ref{tab:results_table}.
% We can see: 
\alias achieves new state-of-the-art (SR: 9.03 vs. 8.57) on the \texttt{test\_unseen} set when there are no semantic and depth labels.
Though counterparts~\citep{episodictransformer,singh2020moca} have better performance on \texttt{test\_seen}, they are likely to be over-fitting by simply memorizing the visible scenes. 
However, our \alias achieves stable results between the \texttt{test\_seen} set and unseen set, demonstrating its generalizability.
In Fig.~\ref{fig:instruction_vis}, we show a trajectory to execute ``place a washed sponge in a tub'' and the intermediate estimates generated by \alias.

When depth supervision is used, our \alias w. depth model has a 17.24\% success rate on the \texttt{test\_unseen} set, as well as competitive goal-condition success rate and path length weighted results.
Note that FILM~\cite{min2021film} leverages additional dense semantic maps as supervision to train a policy network, hence not apple-to-apple comparable to our work. We report the \alias-Oracle model as an upper bound, which learns supervised segmentation and depth, and can be seen as a variant of FILM~\cite{min2021film} without the policy network. It achieves 23.68\% SR on \texttt{test\_unseen}.

\noindent \textbf{Ablation Study.}
We conduct experiments to study the effect of the language representation in concept learning, and the mapping strategy in map construction. 
The results are shown in Fig.~\ref{fig:ablation_concept} and Fig.~\ref{fig:ablation_mapping}, offering 1) benefiting from the natural structure of language, the word embedding is better than the learned encoding, and 2) Bayesian filtering outperforms maximum fusion as the soft probabilities could correct wrong labels.

\begin{figure}[t]
\begin{minipage}[t]{0.49\linewidth}
\begin{table}[H]
\centering
% \fontsize{8}{8}\selectfont
\setlength\tabcolsep{3.7pt} 
\caption{The percentage of failure cases belonging to different failure modes on validation set.}
% \vspace{-0.3em}
\renewcommand\arraystretch{1.1}
\resizebox{1\linewidth}{!}{
\begin{tabular}{@{}lccc@{}}
\shline
Error mode
   &\multicolumn{1}{c}{Seen \%} && \multicolumn{1}{c}{Unseen \%} \\
\shline
% \;Target not found              &9.28   &&12.88  \\    %
Grounding error/Target not found              &36.38   &&28.53  \\    %
Interaction failures          &6.59   &&10.39\\    %
Collisions                    &4.34   &&4.43\\    %
Blocking/Object not accessible &31.29 &&39.75  \\    %
% \;Exceed step limit             &27.10  &&15.65\\
% \;Language processing error  \\    %
Others                        &21.41  &&16.90\\    %
\shline
\end{tabular}}
\label{tab:error_modes}
\end{table}
\end{minipage}
\hfill
\begin{minipage}[t]{0.49\linewidth}
\begin{table}[H]
\footnotesize
\centering
\setlength{\tabcolsep}{3pt}
\renewcommand\arraystretch{1.2}
\caption{Concept reasoning accuracy. We leverage \alias to infer whether an object exists or to count its number in a scene.
% learning concept, Reasoning (example, same room, which room)
}
\resizebox{1\linewidth}{!}{
    \begin{tabular}{lccc}
    % \toprule 32
    \shline
    \multirow{1}{*}{Model}  & Grounding \%  & Exist \% & Count \%  \\
    \shline
    Random Guess & -- & 50.0 & 25.0 \\
    C3D~\cite{tran2015learning}  & -- & 78.1 & 34.4 \\
    \alias (Ours) & 57.6  & 90.6 &  56.3 \\
    \shline
    \end{tabular}
    }
    \label{tab:count}
\end{table}
\end{minipage}
\vspace{-12pt}
\end{figure}

\subsection{Evaluation of Concept Learning}
% In this section, we quantitatively evaluate our concept learning module and show visualization results and error modes.

\noindent \textbf{Quantitative Evaluation.} We report the per-task evaluation results in Fig.~\ref{fig:concept_small}. The concept learning accuracies of objects ``HandTowel'', ``Laptop'', ``Bowl'', and ``Knife'' are above 80\%, because these objects frequently appear alone in the scene (easy to learn and less likely to be confused).
Objects like ``Basketball'', ``Glassbottle'', ``Cellphone'', and ``Teddybear'' are rarely shown in the environment, thus their concepts are difficult to learn.
We also notice that the object ``apple'' appears very rarely, but our model grounds its concept well with the help of language embeddings, \eg, the relationship between ``tomato'' and ``apple''.

\noindent \textbf{Error Modes.} Tab.~\ref{tab:error_modes} shows the error mode of \alias w. depth on ALFRED validation set. We see that ``blocking and object not accessible'' is the most common error mode, which is mainly caused by incorrectly estimated depth or undetected visual objects/concepts.
Additionally, around 30\% of the failures are due to wrongly grounded concepts or the target object not being found.
If we replace our unsupervised concept learning with supervised semantics (\alias-Oracle), the percentage of the error mode for ``Grounding error/Target not found'' changes to 7.38\% and ``blocking and object not accessible'' becomes 44.00\%.

\noindent \textbf{Visualization.} We visualize our concept learning results in Fig.~\ref{fig:concept_vis} by showing the original image, the supervised learned semantics, and our grounded semantics by the concept learner.
We observe our concept learning keeps more object proposals than the supervised model. While most of the main objects in an image can be grounded correctly, there exist a few wrong labels in overlapped or corner areas.
We also show two failure cases on the third and fourth rows of Fig.~\ref{fig:concept_vis}. The first one recognizes ``floor'' as ``diningtable'', a bug that could be fixed by our Bayesian filtering-based semantic mapping.
The other one identifies ``coffeetable'' as ``drawer'', which causes the error ``target not found''. The instruction would succeed if we take the ground truth concept for ``coffeetable''.

\input{figures/ground_vis}

\subsection{Concept Reasoning}
In addition to EIF, we show the learned concept can be transferred to embodied reasoning tasks, \eg, (i) the existence of objects in the scene, (ii) count the number of objects in the scene (Fig.~\ref{fig:qa}).
We build the reasoning dataset by randomly sampling 16 objects from 10 scenes, of which 8 scenes are used for training and the other 2 for testing. A na\"ive baseline is random guessing with 50\% accuracy for the exist task and 25\% accuracy for the count task.
We also train a C3D model~\cite{tran2015learning} that samples 16 frames as input and outputs predictions directly.
Our \alias performs clear and step-by-step interpretable reasoning through semantic grounding and mapping. 
As Tab.~\ref{tab:count} shows, it outperforms both baselines by a large margin. 
By embodied concept learning, \alias can also resolve high-level 3D question-answering tasks, like ``whether two objects appear on a table'' in Fig.~\ref{fig:qa}.
% and ``what is the relative position of the two objects?''.

\section{Discussions}
% \section{Discussion and Limitations}
% \my{performance bottlenecks: depth}
% Explain the reason why we still use supervised depth. Gaps.
This paper proposes \alias, a general framework that can ground visual concepts, build semantic maps and plan actions to accomplish tasks by learning purely from human demonstrations and language instructions.
% , without access to ground-truth semantic and depth supervisions from simulations.
While achieving good performance on embodied instruction following, \alias has limitations.
% It currently focuses solely on learning object concepts and 3D layouts through interactive environments. It would be exciting to extend the framework to learn more dynamic action concepts (\eg ``cutting tomatos'' and ``picking up a knife'') and apply them to more diverse downstream tasks like action grounding and retrieval~\cite{caba2015activitynet,zhou2019grounded}.  
%\my{only learn object concepts, ground actions,  retrieval in the future.}
Although the ALFRED benchmark is photo-realistic, comprehensive, and challenging, there still exists a gap between the embodied environment and the real world. We leave the physical deployment of the framework as our future work. 
%\my{ALFRED is photo-realistic and challenging, still exists gaps between ,we will try Physical deployment in the future}
% The proposed approach has no ethical or societal issues on its own, except those inherited from robotics.

\noindent \textbf{Acknowledgements.} This work is supported by MIT-IBM Watson AI Lab and its member company Nexplore, Amazon Research Award, ONR
MURI, DARPA Machine Common Sense program, ONR (N00014-18-1-2847), and Mitsubishi
Electric.
Ping Luo is supported by the General Research Fund of HK No.27208720, No.17212120, and No.17200622.

% no \bibliographystyle is required, since the corl style is automatically used.
\bibliography{corl}  % .bib

\newpage

\appendix

\input{figures/ground_vis_supp}

\noindent {\textbf{\Large Appendix}}

\section{ALFRED Dataset}
We evaluate our method and its counterparts on ALFRED~\cite{shridhar2020alfred}, which is a benchmark for connecting human language to actions, behaviors, and objects in interactive visual environments. Planner-based expert
demonstrations are accompanied by both high- and low-level human language instructions in 120 indoor scenes in
AI2-THOR 2.0~\cite{ai2thor}.
ALFRED~\cite{shridhar2020alfred} includes 25,743 English language directives describing 8,055 expert demonstrations averaging 50 steps each, resulting in 428,322 image-action pairs.
The test set contains ``Test seen'' (1,533 episodes) and ``Test unseen" (1,529 episodes); the scenes of the latter entirely consist of rooms that do not appear in the training set, while those of the former only consist of scenes seen during training. Similarly, the validation set contains ``Valid seen'' (820 episodes) and ``Valid Unseen" (821 episodes). The success rate is the ranking metric used in the official leaderboard.

\section{Evaluation of the Semantic Policy}
In this section, we evaluate the semantic policy used in our model (average semantic probability map from demonstrations), in comparison to the learned semantic policy in FILM~\cite{min2021film}.
\cite{min2021film} learns a semantic policy model using additional map supervision. However, the policy in our work is freely available from the grounded average semantic probability map.
From Tab.~\ref{tab:policy}, we can see that: our average semantic probability map achieves good performance. With the learned semantic policy, the success rate improves slightly from 17.24\% to 17.92\%.
To keep our framework clean with reduced supervision, we train our model without learning a semantic policy~\cite{min2021film}.

\input{tables/policy_compare}

\input{tables/per-task-performance}

\input{tables/grounding_acc_small}

\input{tables/grounding_acc_large}

\begin{figure}[t]
    \centering
\includegraphics[width=0.95\linewidth]{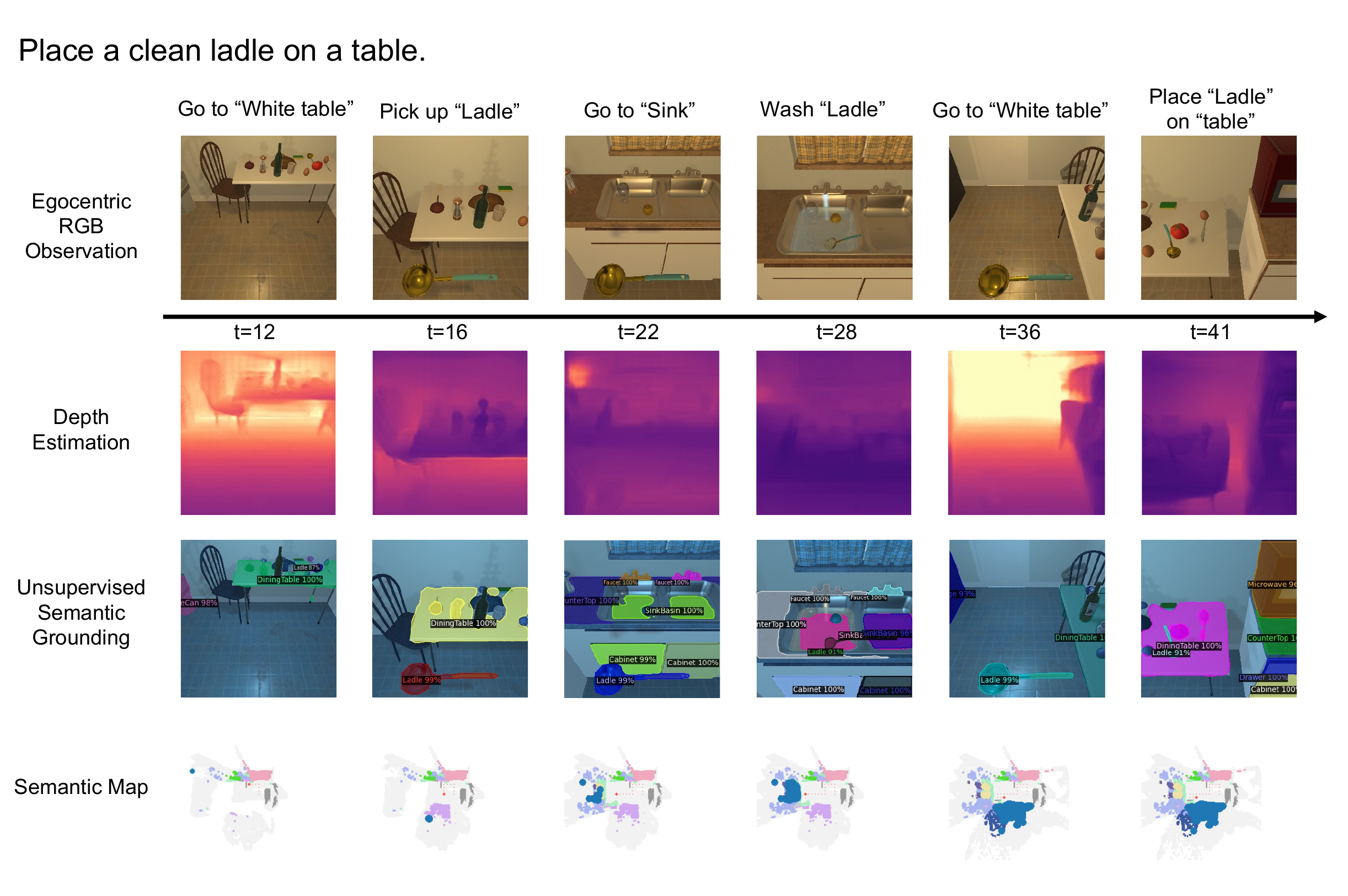}
\includegraphics[width=0.95\linewidth]{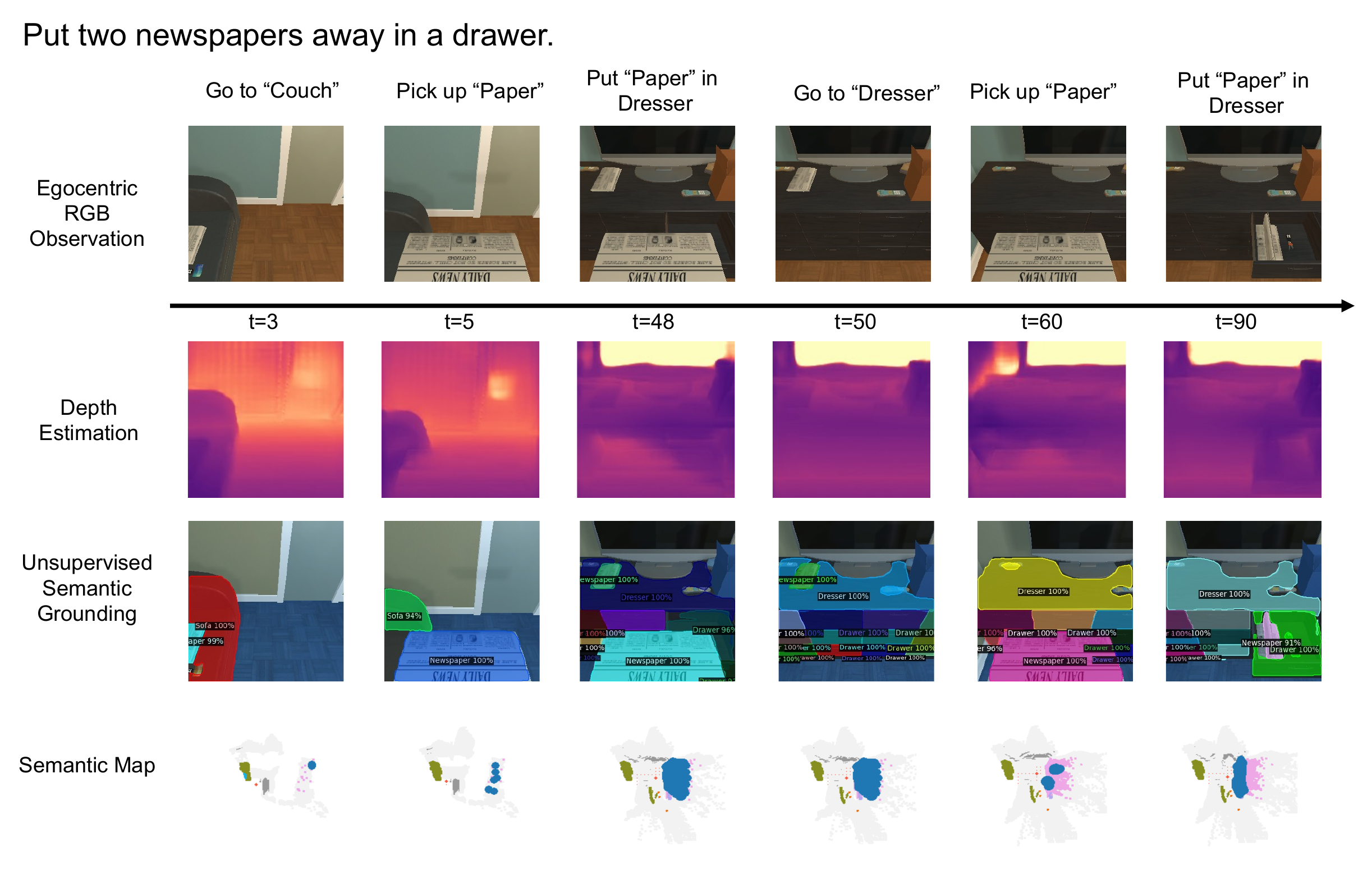}
    \vspace{-6pt}
    \caption{Two examples of intermediate estimates for \alias when the agent tries to accomplish the instructions. Based on RGB observations, our system estimates the depths and semantic masks. The BEV semantic map is gradually established with these estimates as exploration continues.
    The goals (sub-goal/final-goal) are represented by large blue dots in the semantic map, while the agent trajectories are plotted as small red dots.
    }
    % \vspace{-18pt}
    \label{fig:instruction_vis_supp}
\end{figure}

% \begin{figure}[t]
%     \centering
% % \includegraphics[width=0.99\linewidth]{figures/sup/vis/vis1_yx.pdf}
% \includegraphics[width=0.99\linewidth]{figures/sup/vis/vis2_yx.pdf}
%     \vspace{-6pt}
%     \caption{Another example of intermediate estimates by \alias when an agent tries to accomplish an instruction. Based on the RGB observations, our system estimates the depths and semantic masks. The BEV semantic map is gradually established with these estimates as the exploration goes on.
%     The goals (sub goal/final goal) are represented by big blue dots in the semantic map, while the agent trajectories are plotted as small red dots.
%     }
%     % \vspace{-18pt}
%     \label{fig:instruction_vis_supp}
% \end{figure}

\section{Per-task Performance} 
We provide per-task performance (success rate and goal-condition success rate) in Tab.~\ref{tab:ablations_by_type} to show \alias's strengths and weaknesses in different types of tasks. 
We have the following observations:
1) ``Stack \& Place'' and ``Cool \& and Place'' are the most challenging tasks, with a low success rate. 
2) The ``Examine'' task is the easiest task, with a success rate over 30\% and 46.81\% goal-condition success rate.
3) A similar observation with FILM~\cite{min2021film} regarding the number of subtasks and success rate is found: whereas ``Heat \& Place'' and ``Clean \& Place'' usually involve three more subtasks than ``Pick \& Place'', the metrics of the former are higher than the latter. 
This is because ``Heat \& Place'' only appears in kitchens, and ``Clean \& Place'' only appears in toilets. And the room area of these two scenes is relatively small. The results show that the success of a task is highly dependent on the type and scale of the scene.

\section{Detailed Analysis of Concept Learning}
In addition to the figure shown in our main paper, we also report the per-object concept grounding evaluation results (small) in Tab.~\ref{tab:grounding_small}. Objects ``HandTowel'', ``KeyChain'', ``Bowl'', and ``Television'' have over 80\% concept learning accuracy because these objects often appear alone in the scene (easy to learn and less likely to be confused).
Objects like ``HandTowel'', ``KeyChain'', ``Bowl'', and ``Television'' rarely appear in the environment, so their concepts are difficult to learn.

Likewise, we perform detailed evaluations and report the per-object concept grounding evaluation results (large) in Tab.~\ref{tab:grounding_large}.
We notice two classes with ground accuracy of 0: ``BathtubBasin'' and ``Desk''. This is because all BathtubBasins are identified as SinkBasins by our grounding model (SinkBasins are shown more frequently than BathtubBasins).
As for ``Desk'',  the object proposals are identified as ``CoffeTable'', ``DiningTable'', etc.
The average grounding accuracy for large objects is higher than for small objects, because large objects often appear alone in the scene (easy to learn and less confusing).

\section{Details of the deterministic policy}
The deterministic policy is based on the Fast Marching Method [56]. If the object needed in the current subtask is observed in the current semantic map, the location of the object is selected as the goal; otherwise, we sample the location based on the distribution of the corresponding object class in our averaged semantic map as the goal. In both cases, we did not use any domain knowledge about ALFRED. We find the goal from the concept learner and plan the shortest path to the goal based on our semantic map. It's a very general solution that can be used in many other tasks or environments rather than a hand-coded policy for ALFRED.

\section{More Visualization Results}
In this section, we show more visualizations of our concept learning results in Figure~\ref{fig:concept_vis_supp}.
We also demonstrate two examples of intermediate estimates by \alias when an agent tries to accomplish an instruction in Figure~\ref{fig:instruction_vis_supp}.

% \begin{figure}[t]
%     \centering
% % \includegraphics[width=0.95\linewidth, trim=0cm 1cm 9cm 1cm, clip]{figures/instruction_vis/res_vis.pdf}
% \includegraphics[width=0.99\linewidth]{figures/concept_freq.pdf}
%     \vspace{-6pt}
%     \caption{\textcolor{red}{We report the exposure frequency v.s. concept learning accuracy of object classes. The Spearman correlation coefficient is 0.58, indicating that they are positively correlated. The figure of the comparison is attached. We can see that the two ranks are highly related, though some outliers exist. For example, the "knife" category appears most frequently, but the grounding accuracy is not high, because it always appears together with other categories like "fork", leading to confusion in the concept learner.}
%     }
%     \vspace{-18pt}
%     \label{fig:concept_freq}
% \end{figure}

\section{Key Assumption for Other Simulators or Real Robots}
Real robotics applications have been one of the longstanding motivations for this work and have been carefully considered by the authors in the design of ALFRED. When generalizing our model to real-world scenarios:
\begin{itemize}
\vspace{-6pt}
    \item  The instruction parser is supervisedly trained, and can be directly employed in the real world.
    \item  The embodied concept learner should work when there are real-world demonstrations. Currently, we have some assumptions based on the artifact of the AI2THOR environment. However, the assumptions are not strong and are still applied to real environments.
    \item  Unsupervised depth and mapping are well-studied problems in the real world. We see this as a reasonable assumption for the time being. 
    \item  Still, there are some limitations in ALFRED that the action execution is not feasible, i.e, picking up an object by only one command without robot manipulation. However, the low-level control task and the current embodied instruction following task are orthogonal, which means the two tasks can still be decoupled in real-world scenarios, while our model focus on instruction following. 
\end{itemize}
However, it's really challenging for a robot to perform instruction following in an unseen real-world environment, even in a simulated environment (test unseen success rate of only 23.6\% even in our oracle model). To this end, ALFRED simplifies the hard problem of making meaningful progress through tight integration between visual perception, language instruction, and robotic navigation and manipulation. To the best of our knowledge, no other benchmarks contain language instructions in an interactive 3D environment with visual observation and navigation. As the field progresses, we are confident more works and benchmarks will be introduced, and we will take it as our future research direction.

\section{Related Work about Depth and Mapping}
Depth estimation~\cite{hui2022rm,luo2020consistent,ranjan2019competitive,gordon2019depth,zhou2017unsupervised,godard2019digging,yao2018mvsnet,lu2021global} has witnessed a boom since the emergence of deep learning. 
Compared with stereo matching~\cite{yao2018mvsnet,guo2019group,wang2021patchmatchnet} and sensor-based  methods~\cite{tang2020learning,xu2019depth}, the monocular depth estimation only requires a single-view color image for depth inference, which is suitable for practical deployment given its low-cost nature. 
Following the supervised methods~\cite{laina2016deeper,cao2017estimating}, Zhou et al.~\cite{zhou2017unsupervised} first demonstrated the possibility of depth learning in an unsupervised manner, inspired by the learning principle of humans. Afterwards, the unsupervised depth estimation are well explored in both indoor~\cite{yang2020d3vo,li2021structdepth,ji2021monoindoor} and outdoor scenarios~\cite{godard2019digging,pillai2019superdepth,hui2022rm} due to its labeling-free advantage. In this work, we also follow their spirits to investigate the learning process of an agent baby.
After depth estimation, a mapping module~\cite{mur2015orb,izadi2011kinectfusion,shan2018lego} is usually included in a robotic system to memorize the geometry layouts of the visited regions for path planning and navigation. Given different sensor properties and map representations, the mapping procedure could also differ. For instance, \cite{klein2007parallel,mur2015orb} maintain reliable sparse landmarks, \cite{izadi2011kinectfusion} constructs TSDF, and \cite{zhang2014loam,shan2018lego} store voxel maps.

\end{document}

%% file: tables/main_tbl.tex
\begin{table}[t]
\centering
% \fontsize{8}{8}\selectfont
\setlength\tabcolsep{8pt} 
\renewcommand\arraystretch{1}
\caption{
% Test results on the ALFRED benchmark. 
Comparison with other methods on ALFRED benchmark. 
The 
% top section 
upper part 
contains unsupervised methods while the lower part contains the supervised counterparts with semantic or depth supervisions. 
% methods in the bottom section leverage semantic or depth supervision.
% Note that FILM~\cite{min2021film} leverage additional dense semantic maps as supervision to train a policy network, hence not apple-to-apple comparable to our work. 
We also report the \alias-Oracle model as an upper bound,
% which learns 
with supervised segmentation and depth.  
% which can be seen as a variant of FILM~\cite{min2021film} without the policy network.
The top scores are in \textbf{bold}. 
% \textbf{Bold} numbers are top scores in each section. 
\textcolor{MyDarkRed}{\textbf{Red}} denotes the top success rate (SR) (ranking metric of the leaderboard) on the \texttt{test\_unseen} set. 
% (the major ranking metric of the leaderboard)
% (by which the leaderboard is ranked).
% `*' indicates the work uses learned depth to build a 3D or bird's eye view map.
}
% \begin{tabular}{@{}lr@{\hspace{7pt}}c@{\hspace{7pt}}c@{\hspace{7pt}}ccc@{\hspace{7pt}}c@{\hspace{7pt}}c@{\hspace{7pt}}c@{}}
\resizebox{\linewidth}{!}{
\begin{tabular}{l c c c c c c c c c c c c }
\toprule
\multirow{3}{*}{\textbf{Method}} &  \multicolumn{2}{c}{\textbf{Supervision}}  && \multicolumn{4}{c}{\textbf{Test Seen}} && \multicolumn{4}{c}{\textbf{Test Unseen}} \rowsqueeze \\
 \cmidrule{2-3} \cmidrule{5-8}\cmidrule{10-13}\rowsqueeze
  & Semantic & \makecell{Depth} && \makecell{PLWGC \\ (\%)} &  \makecell{GC \\ (\%)}  & \makecell{PLWSR \\ (\%)} & \makecell{SR \\ (\%)} && \makecell{PLWGC \\ (\%)} & \makecell{GC \\ (\%)}  & \makecell{PLWSR \\ (\%)} & \makecell{\textbf{\textcolor{MyDarkRed}{SR}} \\ \textbf{\textcolor{MyDarkRed}{(\%)}}} \\
\midrule
% \multicolumn{10}{l}{\textbf{Supervised Learning (Semantic Labels Provided)}} \rowsqueeze\\
% \midrule
\textsc{Seq2Seq}~\citep{shridhar2020alfred}    &  $\times$ & $\times$
             &&  6.27   & 9.42  & 2.02  &   3.98   
                  && 4.26    & 7.03  & 0.08   & 3.90 \\ 

\textsc{MOCA}~\citep{singh2020moca}     & $\times$ & $\times$
                 && 22.05   & 28.29 &  15.10  &    22.05   
                  && 9.99   & 14.28   & 2.72   & 5.30 \\  

\textsc{LAV}~\citep{lav}    & $\times$ & $\times$
               && 13.18    &  23.21 &   6.31  &   13.35   
                  && 10.47    &17.27  & 3.12   & 6.38 \\    
% \textsc{E.T.}~\citep{episodictransformer}    & & &
%                  && -  &  36.47  & -     &  28.77   
%                   && - & 15.01    & -   & 5.04 \\  
\textsc{E.T.}~\citep{episodictransformer}   & $\times$ & $\times$
                 && \textbf{34.93}   &  \textbf{45.44}  &  \textbf{27.78}   &  \textbf{38.42}   
                  && 11.46   & 18.56   & 4.10    & 8.57 \\
% \textsc{HiTUT}~\citep{hitut}    &  &  &
%                     &&  17.41  & 29.97   & 11.10  &    21.27    
%                   && 11.51   & 20.31    &  5.86   & 13.87 \\    

% \rowcolor{Gray}
% \textsc{\alias-Oracle (Ours, old score)} & $\surd$ & & $\surd$ 
% && 14.42 & 34.89 & 9.95 & 23.42 
% && 12.67 & 31.95 &  8.95 & 20.21 \\
% \rowcolor{Gray}
% \textsc{\alias-Oracle (Ours, new score)} & $\surd$ & & $\surd$ 
% && 12.74 & 27.98 & 8.67 & 18.79
% && 11.52 & 27.75 & 7.45 & 17.92 \\
\rowcolor{Gray}
\textsc{\alias (Ours)} & $\times$ &  $\times$
&&9.47  &18.74  &4.97  & 10.37
&& \textbf{11.50} & \textbf{19.51} & \textbf{4.13} & \color{MyDarkRed}{\textbf{9.03}} \\
% \rowcolor{Gray}
% \textsc{\alias (Ours)} & $\times$ &  $\times$
% &&9.47  &18.74  &4.97  & 10.37
% && \textbf{12.13} & \textbf{19.18} & 4.09 & \color{MyDarkRed}{\textbf{9.03}} \\
\hline
\textsc{EmBERT}~\cite{suglia2021embodied} & $\surd$ & $\times$
                &&  \textbf{32.63}   & 38.40  &  24.36  &  31.48   
                  && 8.87   & 12.91   &  2.17   & 5.05 \\   
\textsc{LWIT}~\citep{lwit}   & $\surd$ & $\times$
                &&  23.10   & 40.53  &  \textbf{43.10}  &  30.92   
                  && \textbf{16.34}   & 20.91   &  5.60   & 9.42 \\   
                  
% \textsc{HiTUT} G-only~\citep{hitut}    & & &
%               &&  -  &  21.11  &  - &  13.63     
%                   && -   & 17.89  & -  & 11.12 \\    
\textsc{HiTUT}~\citep{hitut}    & $\surd$  & $\times$
                    &&  17.41  & 29.97   & 11.10  &    21.27    
                  && 11.51   & 20.31    &  5.86   & 13.87 \\    

\textsc{ABP}~\citep{abp}    & $\surd$ & $\times$
                && 4.92  &  \textbf{51.13}   & 3.88   &  \textbf{44.55}     
                  && 2.22   & 24.76   & 1.08   & 15.43  \\   

\textsc{VLNBERT}~\citep{song2022one} & $\surd$ & $\times$
 && 19.48 & 33.35 & 13.88 & 24.79
 && 13.18 & 22.60 & 7.66 & 16.29\\ 

\textsc{HLSM}~\citep{blukis2021persistent}    & $\surd$ & $\surd$
               &&  11.53  &  35.79   &  6.69 &  25.11   
                  && 8.45  &27.24    & 4.34   & 16.29 \\      

% \textsc{FILM*}~\cite{min2021film}& $\surd$ & $\surd$
%              &&  15.06  &38.51   & 11.23   &    27.67    
%                   && \textbf{14.30}    & \textbf{36.37}    &  \textbf{10.55}    & \color{red}{\textbf{26.49}} \\
                  
% \textsc{FILM w.o. Semantic Search*}~\cite{min2021film}  & $\surd$ & 
%               && 13.10 & 35.59   & 9.43   &   25.90     
%                   && 13.37   & 35.51   & 10.17  & 23.94  \\   

\rowcolor{Gray}
\textsc{\alias w. depth (Ours)} & $\times$ & $\surd$
&& 12.34 & 27.86 & 8.02 & 18.26
&& 11.11 & \textbf{27.30} & \textbf{7.30} & \color{MyDarkRed}{\textbf{17.24}} \\
\midrule
\rowcolor{Gray}                  
\textsc{\alias-Oracle (Ours)} & $\surd$ & $\surd$
&& 15.19 & 36.40 & 10.56 & 25.90
&& 13.08 & 35.02 & 9.33 & 23.68 \\
% \textsc{\alias-Oracle (Ours)} & $\times$ & GT 
% && 15.19 & 36.40 & 10.56 & 25.90
% && 13.08 & 35.02 & 9.33 & 23.68 \\
% \textsc{\alias-Oracle (Ours)} & GT & $\times$
% && 15.19 & 36.40 & 10.56 & 25.90
% && 13.08 & 35.02 & 9.33 & 23.68 \\
% \textsc{FILM*}~\cite{min2021film}& $\surd$ & $\surd$
%              &&  15.06  &38.51   & 11.23   &    27.67    
%                   && \textbf{14.30}    & \textbf{36.37}    &  \textbf{10.55}    & \color{MyDarkRed}{\textbf{26.49}} \\
% \multicolumn{10}{l}{\textbf{Unsupervised Learning (Instructions Only)}} \rowsqueeze \\
% \midrule
% \rowcolor{Gray}
% \textsc{\alias (Ours)} \my{xuyan} &  & & &  \\
\bottomrule
\end{tabular}}
\label{tab:results_table}
\vspace{-12pt}
\end{table}

%% file: figures/ground_vis.tex
\begin{figure}[t]
\begin{minipage}[t]{0.343\linewidth}
\begin{figure}[H]
    \centering
    \includegraphics[width=\linewidth]{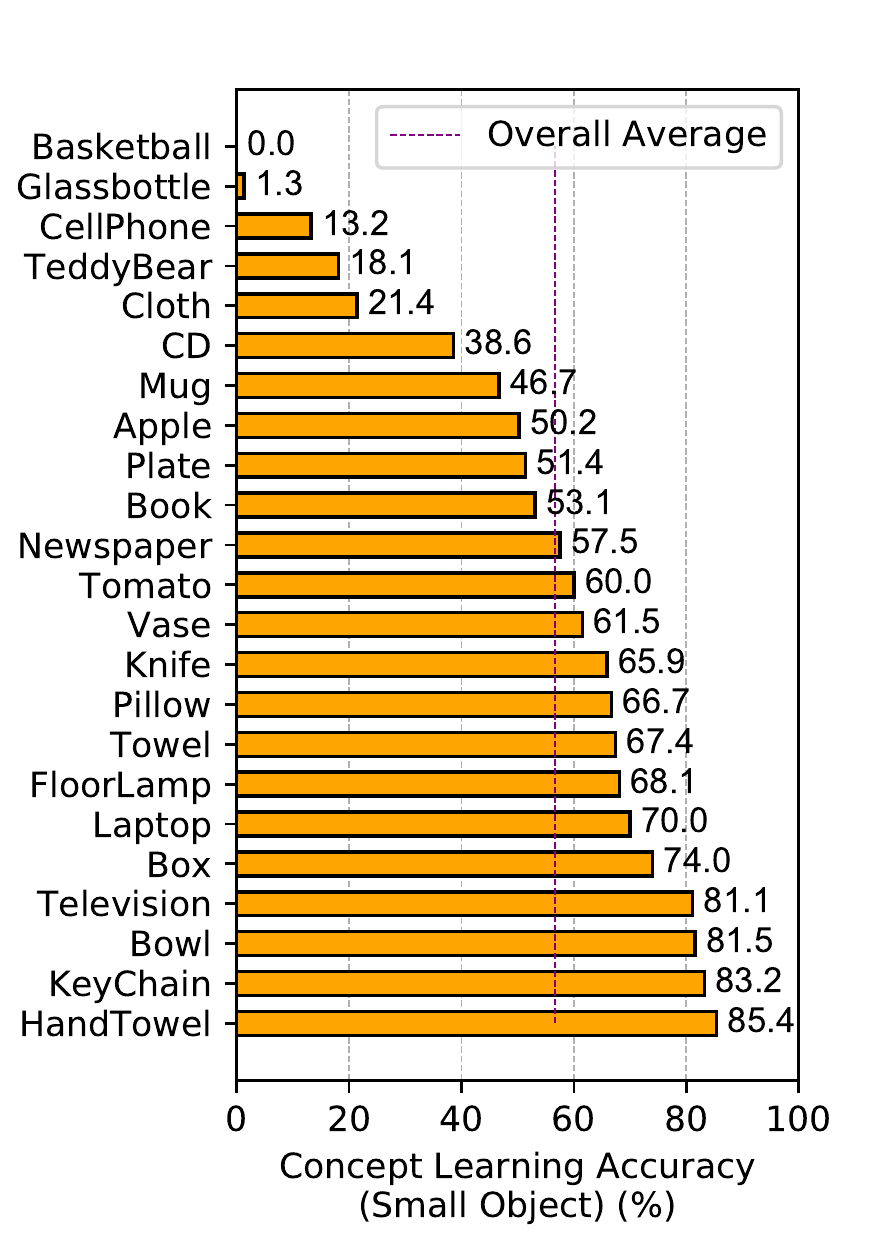}
    \vspace{-17pt}
    \caption{
    % Object grounding accuracy with our concept learning.
    Concept learning accuracy.
    % on small objects. 
    % Per-class concept learning accuracy. 
    % (ordered). 
    Results for challenging small objects are shown. 
    % Please refer to the appendix for 
    Complete analyses are in appendix. 
    % More results are included in the Appendix. 
    % and receptacle objects are shown in Appendix. 
    % We show 23 small objects 
    % % (could be picked up) 
    % and the overall accuracy here, more results and the receptacle objects are shown in Appendix.
    }
   \label{fig:concept_small}
    % \vspace{-20pt}
    % \vspace{-9ex}
\end{figure}
\end{minipage}
\hfill
\begin{minipage}[t]{0.298\linewidth} % 37
\begin{figure}[H]
    \centering
    \includegraphics[width=\linewidth]{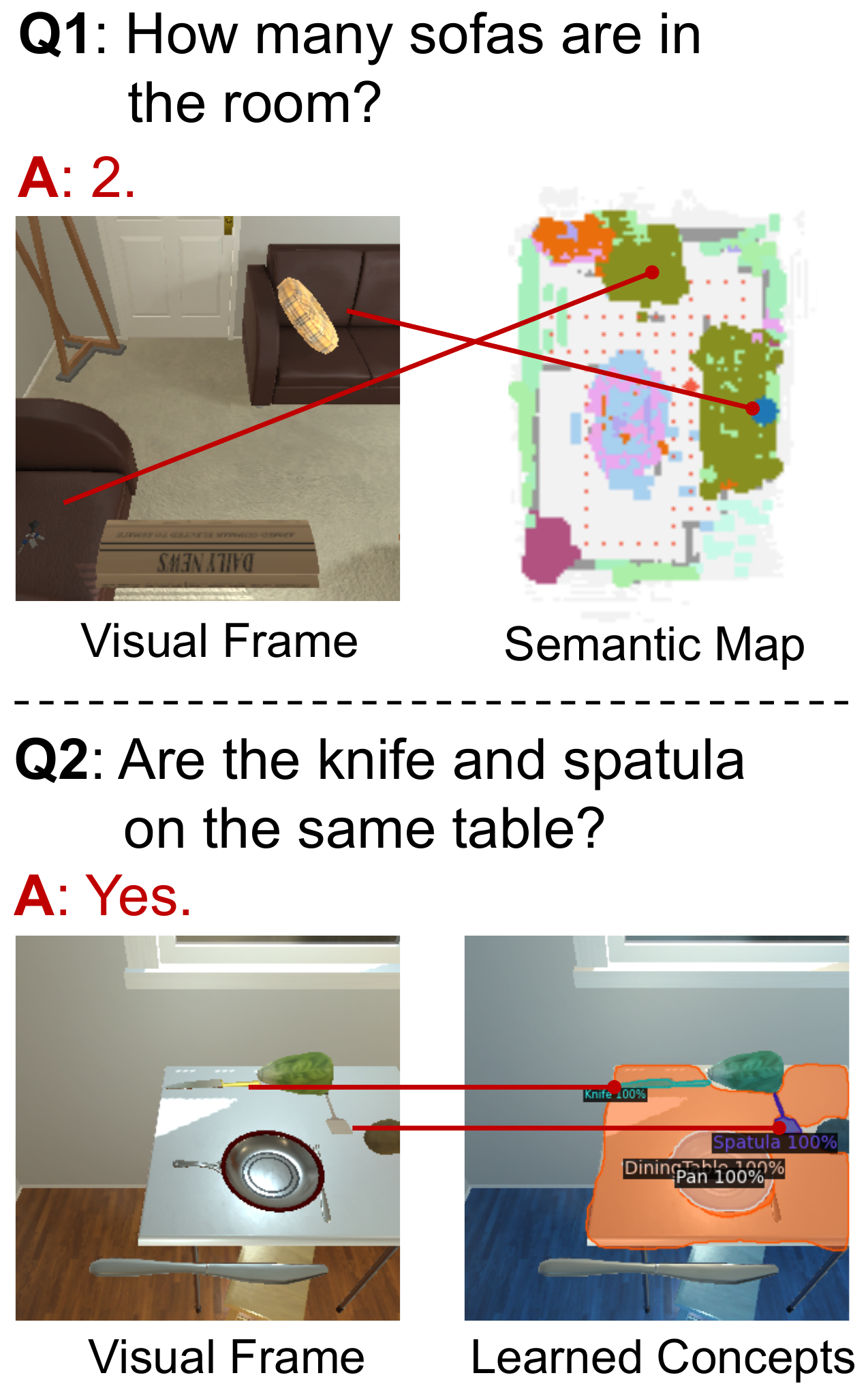}
    \vspace{-17pt}
    \caption{Examples of concept reasoning by \alias:
    the count task and the high-level question-answering.
    }
    \label{fig:qa}
\end{figure}
\end{minipage}
\hfill
\begin{minipage}[t]{0.328\linewidth} % 61
\begin{figure}[H]
% \begin{wrapfigure}{R}{0.54\textwidth}
\centering
\footnotesize
% \vspace{-42pt}
\includegraphics[width=.32\columnwidth]{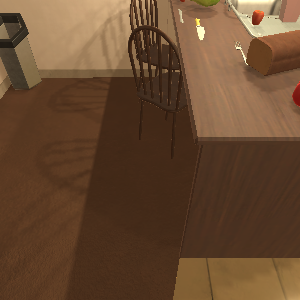}~
\includegraphics[width=.32\columnwidth]{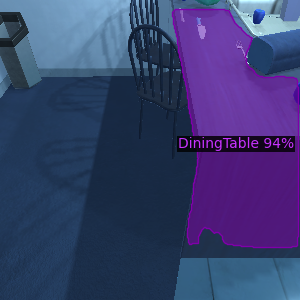}~
\includegraphics[width=.32\columnwidth]{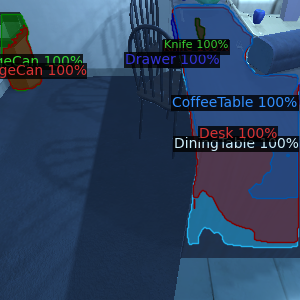}~\\
\vspace{1mm}
\includegraphics[width=.32\columnwidth]{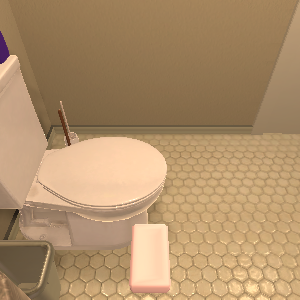}~
\includegraphics[width=.32\columnwidth]{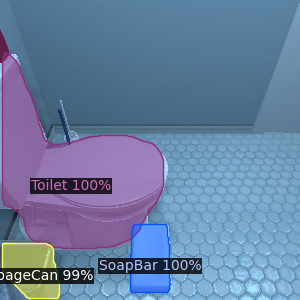}~
\includegraphics[width=.32\columnwidth]{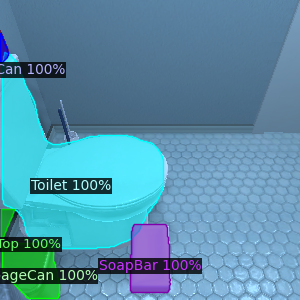}~\\
\vspace{1mm}
\includegraphics[width=.32\columnwidth]{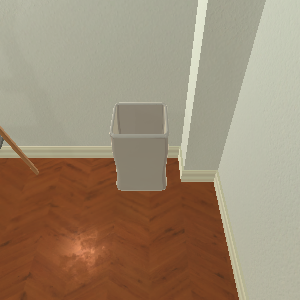}~
\includegraphics[width=.32\columnwidth]{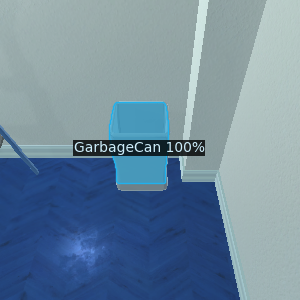}~
\includegraphics[width=.32\columnwidth]{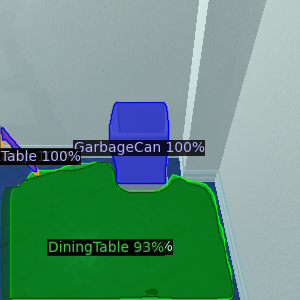}~\\
\vspace{1mm}
\includegraphics[width=.32\columnwidth]{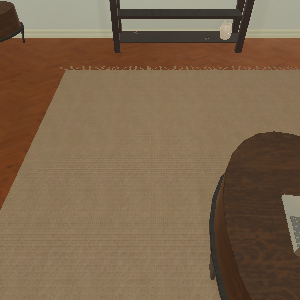}~
\includegraphics[width=.32\columnwidth]{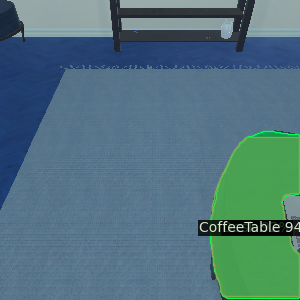}~
\includegraphics[width=.32\columnwidth]{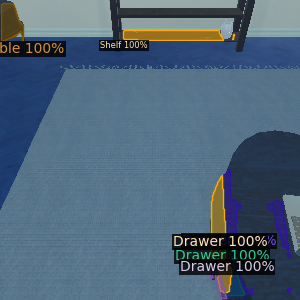}~\\
\vspace{-4pt}
\caption{Concept learning visualization.
 From left to right: the original image, supervised instance segmentation map, and our concept learning results.
}
\label{fig:concept_vis}
\end{figure}
% \end{wrapfigure}
\end{minipage}
\vspace{-12pt}
\end{figure}

%% file: figures/ground_vis_supp.tex
\begin{figure}[t]
% \begin{wrapfigure}{R}{0.54\textwidth}
\centering
\footnotesize
% \vspace{-42pt}
\includegraphics[width=.19\columnwidth]{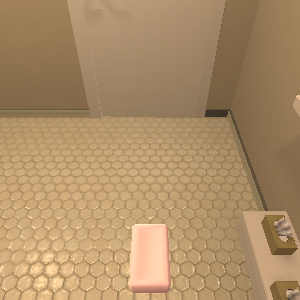}~
\includegraphics[width=.19\columnwidth]{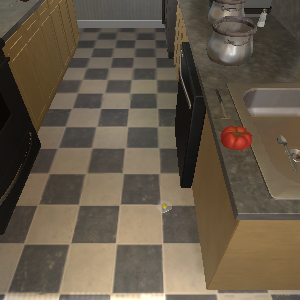}~
\includegraphics[width=.19\columnwidth]{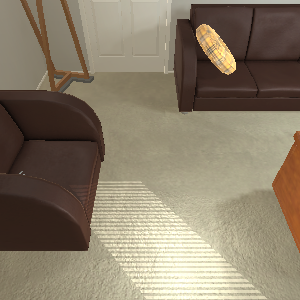}~
\includegraphics[width=.19\columnwidth]{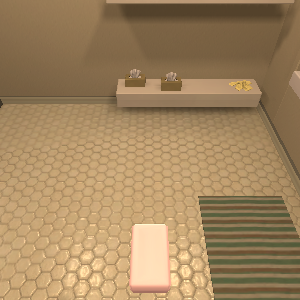}~
\includegraphics[width=.19\columnwidth]{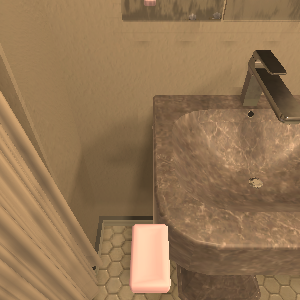}~\\
\vspace{1mm}
\includegraphics[width=.19\columnwidth]{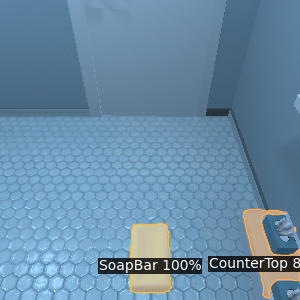}~
\includegraphics[width=.19\columnwidth]{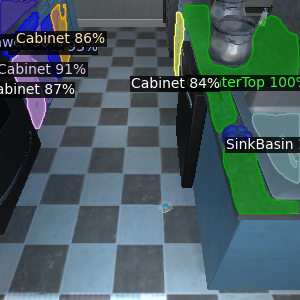}~
\includegraphics[width=.19\columnwidth]{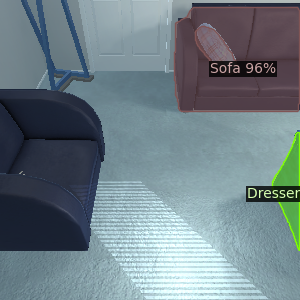}~
\includegraphics[width=.19\columnwidth]{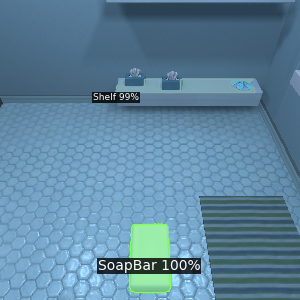}~
\includegraphics[width=.19\columnwidth]{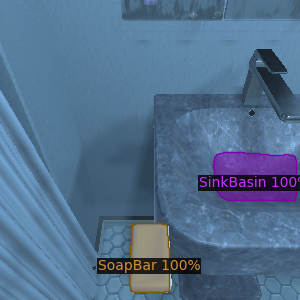}~\\
\vspace{1mm}
\includegraphics[width=.19\columnwidth]{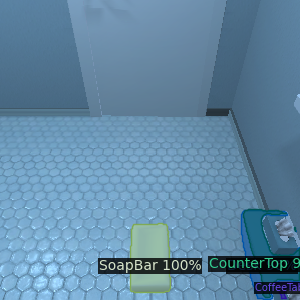}~
\includegraphics[width=.19\columnwidth]{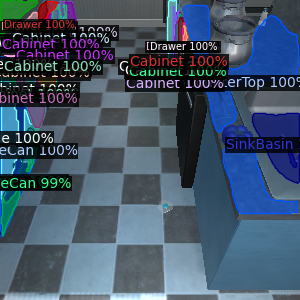}~
\includegraphics[width=.19\columnwidth]{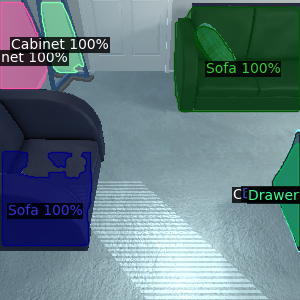}~
\includegraphics[width=.19\columnwidth]{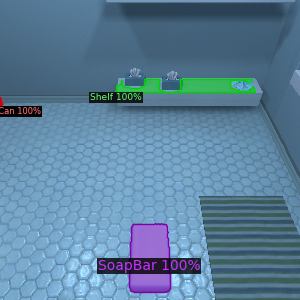}~
\includegraphics[width=.19\columnwidth]{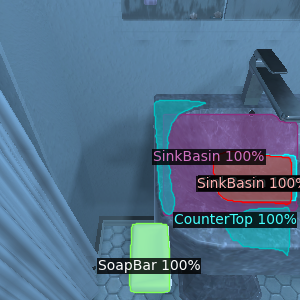}~\\
% \vspace{-4pt}
\caption{Concept learning visualization.
 From top to bottom: the original image, supervised instance segmentation map, and our concept learning results.
}
\label{fig:concept_vis_supp}
\end{figure}

%% file: tables/policy_compare.tex
\begin{table}[t]
\centering
% \fontsize{8}{8}\selectfont
\setlength\tabcolsep{1pt} 
\renewcommand\arraystretch{1.15}
\caption{
% Test results on the ALFRED benchmark. 
Comparison of the semantic policies on the ALFRED benchmark. 
% \textbf{Bold} numbers are top scores in each section. 
\textcolor{MyDarkRed}{\textbf{Red}} denotes the top success rate (SR) (ranking metric of the leaderboard) on the \texttt{test\_unseen} set.
We take our \alias w. depth as the baseline model and make comparison between our model and our model + learned semantic policy~\cite{min2021film}.
% (the major ranking metric of the leaderboard)
% (by which the leaderboard is ranked).
% `*' indicates the work uses learned depth to build a 3D or bird's eye view map.
}
% \begin{tabular}{@{}lr@{\hspace{7pt}}c@{\hspace{7pt}}c@{\hspace{7pt}}ccc@{\hspace{7pt}}c@{\hspace{7pt}}c@{\hspace{7pt}}c@{}}
\resizebox{\linewidth}{!}{
\begin{tabular}{l c c c c c c c c c c c c c }
\toprule
\multirow{3}{*}{\textbf{Method}} &  \multicolumn{3}{c}{\textbf{Supervision}}  && \multicolumn{4}{c}{\textbf{Test Seen}} && \multicolumn{4}{c}{\textbf{Test Unseen}} \rowsqueeze \\
 \cmidrule{2-4} \cmidrule{6-9}\cmidrule{11-14}\rowsqueeze
  & Semantic & \makecell{Depth} & \makecell{Policy} && \makecell{PLWGC \\ (\%)} &  \makecell{GC \\ (\%)}  & \makecell{PLWSR \\ (\%)} & \makecell{SR \\ (\%)} && \makecell{PLWGC \\ (\%)} & \makecell{GC \\ (\%)}  & \makecell{PLWSR \\ (\%)} & \makecell{\textbf{\textcolor{MyDarkRed}{SR}} \\ \textbf{\textcolor{MyDarkRed}{(\%)}}} \\
\midrule
% \rowcolor{Gray}
% \rowcolor{Gray}
\textsc{\alias} & $\times$ & $\surd$ & probability map 
&& 12.34 & 27.86 & 8.02 & 18.26
&& 11.11 & 27.30 & 7.30 & 17.24 \\
\textsc{\alias + Policy} & $\times$ & $\surd$ & learned map
&& 12.74 & 27.98 & 8.67 & 18.79
&& 11.52 & 27.75 & 7.45 & \color{MyDarkRed}{\textbf{17.92}} \\
\bottomrule
\end{tabular}}
\label{tab:policy}
% \vspace{-12pt}
\end{table}

%% file: tables/per-task-performance.tex
\begin{table}[t]
\centering
% \fontsize{8}{8}\selectfont
\renewcommand\arraystretch{1.2}
\centering
% \vspace{10pt}
\captionof{table}{Performance by different task types of model \alias w. Depth on the validation set.}
\footnotesize
\fontsize{8}{8}\selectfont
\setlength\tabcolsep{8pt} 
% \vspace{-0.3em}
\resizebox{1\linewidth}{!}{
\begin{tabular}{@{}lccccc@{}}
\toprule
\multirow{2}{*}{\textbf{Task Type}} &  \multicolumn{2}{c}{\textbf{Val Seen \%}} && \multicolumn{2}{c}{\textbf{Val Unseen \%}} \\
   \cmidrule{2-3} \cmidrule{5-6}
   & Goal-condition & Success Rate  && Goal-condition & Success Rate \\
\midrule
Overall
                & 30.83 &18.67&& 21.74 & 10.50 \\    %
\midrule
Examine
                & 46.81 &31.18    & &47.98&29.65 \\    %
Pick \& Place
                & 21.36 &23.72    & &3.67  &8.49\\    %
Stack \& Place & 16.38  & 6.25    & &7.80  &0.99\\
                % &  {\color{red}how to define} &&&\\    %
Clean \& Place
                & 41.44 &24.77  & &29.50& 8.85    \\    %
Cool \& Place
                & 19.64&5.88  &&13.15&  0.00   \\    %
Heat \& Place
                &35.75 &19.27  &&31.00&13.67 \\    %
Pick 2 \& Place
                &34.48 &19.67 &&19.14& 11.84\\    %
\bottomrule
\end{tabular}
}
\label{tab:ablations_by_type}
\end{table}

%% file: tables/grounding_acc_small.tex
\begin{table}[t]
\footnotesize
\centering
\setlength{\tabcolsep}{1pt}
\renewcommand\arraystretch{1.2}
\caption{Concept grounding accuracy (small).
% Put forward. interpreability, easy to diognise, if we add some data.
}
\resizebox{1\linewidth}{!}{
    \begin{tabular}{l|cccccccc}
    % \toprule
    \shline
    \multirow{1}{*}{Category}   & Vase & Pillow & Plate & Laptop & FloorLamp & Newspaper & HandTowel
    & Box \\
    \hline
    Accuracy (\%) & 61.5 & 66.7 & 51.4 & 70.0 & 68.1 & 57.5 & 85.4 & 74.0 \\
    % \hline
    \shline
    \multirow{1}{*}{Category}   & Towel & Television & Mug & Book & Bowl & Tomato & Knife 
    & KeyChain \\
    \hline
    Accuracy (\%) & 67.4 & 81.1 & 46.7 & 53.1 & 81.5 & 60.0 & 65.9 &  83.2\\
    % \hline
    \shline
    Category & Cloth & TeddyBear & CellPhone & BasketBall & Glassbottle & Apple & CD & Others \\
    \hline
    Accuracy (\%) & 21.4 & 18.1 & 13.2 & 0 & 1.3 & 50.2 & 38.6 & 57.0 \\
    \shline
    \end{tabular}
    }
    \label{tab:grounding_small}
\end{table}
% overall 56.7

%% file: tables/grounding_acc_large.tex
\begin{table}[t]
\footnotesize
\centering
\setlength{\tabcolsep}{1pt}
\renewcommand\arraystretch{1.2}
\caption{Concept grounding accuracy (large). 
}
\resizebox{1\linewidth}{!}{
    \begin{tabular}{l|cccccccccc}
    % \toprule
    \shline
    \multirow{1}{*}{Category}   & Shelf & TVStand & Dresser & Fridge & Microwave & SinkBasin & BathtubBasin  \\
    \hline
    Accuracy (\%) & 77.9 & 82.6 & 75.2 & 13.6 & 64.8 & 99.6 & 0 \\
    % \hline
    \shline
    \multirow{1}{*}{Category}   & CoffeeMachine & Cart & Cabinet & Desk & CoffeeTable & Safe & Drawer \\
    \hline
    Accuracy (\%) & 81.0 & 59.5 & 2.6 & 0 & 74.3 & 73.4 & 38.8 \\
    % \hline
    \shline
    Category & Bed & Sofa & DiningTable & GarbageCan & Toilet & CounterTop  & Others \\
    \hline
    Accuracy (\%) & 64.4 & 72.7 & 52.9 & 53.3 & 81.5 & 87.2 & 59.4 \\
    \shline
    \end{tabular}
    }
    \label{tab:grounding_large}
\end{table}
% overall 58.4

%% file: main.bbl
\begin{thebibliography}{89}
\providecommand{\natexlab}[1]{#1}
\providecommand{\url}[1]{\texttt{#1}}
\expandafter\ifx\csname urlstyle\endcsname\relax
  \providecommand{\doi}[1]{doi: #1}\else
  \providecommand{\doi}{doi: \begingroup \urlstyle{rm}\Url}\fi

\bibitem[Shridhar et~al.(2020)Shridhar, Thomason, Gordon, Bisk, Han, Mottaghi,
  Zettlemoyer, and Fox]{shridhar2020alfred}
M.~Shridhar, J.~Thomason, D.~Gordon, Y.~Bisk, W.~Han, R.~Mottaghi,
  L.~Zettlemoyer, and D.~Fox.
\newblock Alfred: A benchmark for interpreting grounded instructions for
  everyday tasks.
\newblock In \emph{Proceedings of the IEEE/CVF conference on computer vision
  and pattern recognition}, pages 10740--10749, 2020.

\bibitem[Singh et~al.(2020)Singh, Bhambri, Kim, Mottaghi, and
  Choi]{singh2020moca}
K.~P. Singh, S.~Bhambri, B.~Kim, R.~Mottaghi, and J.~Choi.
\newblock Moca: A modular object-centric approach for interactive instruction
  following.
\newblock \emph{arXiv preprint arXiv:2012.03208}, 2020.

\bibitem[Nottingham et~al.(2021)Nottingham, Liang, Shin, Fowlkes, Fox, and
  Singh]{lav}
K.~Nottingham, L.~Liang, D.~Shin, C.~C. Fowlkes, R.~Fox, and S.~Singh.
\newblock Modular framework for visuomotor language grounding.
\newblock \emph{arXiv preprint arXiv:2109.02161}, 2021.

\bibitem[Pashevich et~al.(2021)Pashevich, Schmid, and Sun]{episodictransformer}
A.~Pashevich, C.~Schmid, and C.~Sun.
\newblock Episodic transformer for vision-and-language navigation.
\newblock \emph{arXiv preprint arXiv:2105.06453}, 2021.

\bibitem[Blukis et~al.(2021)Blukis, Paxton, Fox, Garg, and
  Artzi]{blukis2021persistent}
V.~Blukis, C.~Paxton, D.~Fox, A.~Garg, and Y.~Artzi.
\newblock A persistent spatial semantic representation for high-level natural
  language instruction execution.
\newblock In \emph{Proceedings of the Conference on Robot Learning (CoRL)},
  2021.

\bibitem[Min et~al.(2021)Min, Chaplot, Ravikumar, Bisk, and
  Salakhutdinov]{min2021film}
S.~Y. Min, D.~S. Chaplot, P.~Ravikumar, Y.~Bisk, and R.~Salakhutdinov.
\newblock Film: Following instructions in language with modular methods.
\newblock \emph{arXiv preprint arXiv:2110.07342}, 2021.

\bibitem[Walter et~al.(2013)Walter, Hemachandra, Homberg, Tellex, and
  Teller]{walter2013learning}
M.~R. Walter, S.~M. Hemachandra, B.~S. Homberg, S.~Tellex, and S.~Teller.
\newblock Learning semantic maps from natural language descriptions.
\newblock In \emph{Robotics: Science and Systems}, 2013.

\bibitem[Hemachandra et~al.(2015)Hemachandra, Duvallet, Howard, Roy, Stentz,
  and Walter]{hemachandra2015learning}
S.~Hemachandra, F.~Duvallet, T.~M. Howard, N.~Roy, A.~Stentz, and M.~R. Walter.
\newblock Learning models for following natural language directions in unknown
  environments.
\newblock In \emph{2015 IEEE International Conference on Robotics and
  Automation (ICRA)}, pages 5608--5615. IEEE, 2015.

\bibitem[Patki et~al.(2019)Patki, Daniele, Walter, and
  Howard]{patki2019inferring}
S.~Patki, A.~F. Daniele, M.~R. Walter, and T.~M. Howard.
\newblock Inferring compact representations for efficient natural language
  understanding of robot instructions.
\newblock In \emph{2019 International Conference on Robotics and Automation
  (ICRA)}, pages 6926--6933. IEEE, 2019.

\bibitem[Kostavelis and Gasteratos(2015)]{kostavelis2015semantic}
I.~Kostavelis and A.~Gasteratos.
\newblock Semantic mapping for mobile robotics tasks: A survey.
\newblock \emph{Robotics and Autonomous Systems}, 66:\penalty0 86--103, 2015.

\bibitem[Anderson et~al.(2018)Anderson, Wu, Teney, Bruce, Johnson,
  S{\"u}nderhauf, Reid, Gould, and Van Den~Hengel]{anderson2018vision}
P.~Anderson, Q.~Wu, D.~Teney, J.~Bruce, M.~Johnson, N.~S{\"u}nderhauf, I.~Reid,
  S.~Gould, and A.~Van Den~Hengel.
\newblock Vision-and-language navigation: Interpreting visually-grounded
  navigation instructions in real environments.
\newblock In \emph{Proceedings of the IEEE Conference on Computer Vision and
  Pattern Recognition}, pages 3674--3683, 2018.

\bibitem[Fried et~al.(2018)Fried, Hu, Cirik, Rohrbach, Andreas, Morency,
  Berg-Kirkpatrick, Saenko, Klein, and Darrell]{fried2018speaker}
D.~Fried, R.~Hu, V.~Cirik, A.~Rohrbach, J.~Andreas, L.-P. Morency,
  T.~Berg-Kirkpatrick, K.~Saenko, D.~Klein, and T.~Darrell.
\newblock Speaker-follower models for vision-and-language navigation.
\newblock In \emph{Advances in Neural Information Processing Systems}, 2018.

\bibitem[Zhu et~al.(2020)Zhu, Zhu, Chang, and Liang]{zhu2020vision}
F.~Zhu, Y.~Zhu, X.~Chang, and X.~Liang.
\newblock Vision-language navigation with self-supervised auxiliary reasoning
  tasks.
\newblock In \emph{Proceedings of the IEEE/CVF Conference on Computer Vision
  and Pattern Recognition}, pages 10012--10022, 2020.

\bibitem[Ke et~al.(2019)Ke, Li, Bisk, Holtzman, Gan, Liu, Gao, Choi, and
  Srinivasa]{ke2019tactical}
L.~Ke, X.~Li, Y.~Bisk, A.~Holtzman, Z.~Gan, J.~Liu, J.~Gao, Y.~Choi, and
  S.~Srinivasa.
\newblock Tactical rewind: Self-correction via backtracking in
  vision-and-language navigation.
\newblock In \emph{Proceedings of the IEEE/CVF Conference on Computer Vision
  and Pattern Recognition}, pages 6741--6749, 2019.

\bibitem[Wang et~al.(2019)Wang, Huang, Celikyilmaz, Gao, Shen, Wang, Wang, and
  Zhang]{wang2019reinforced}
X.~Wang, Q.~Huang, A.~Celikyilmaz, J.~Gao, D.~Shen, Y.-F. Wang, W.~Y. Wang, and
  L.~Zhang.
\newblock Reinforced cross-modal matching and self-supervised imitation
  learning for vision-language navigation.
\newblock In \emph{Proceedings of the IEEE/CVF Conference on Computer Vision
  and Pattern Recognition}, pages 6629--6638, 2019.

\bibitem[Ma et~al.(2019)Ma, Wu, AlRegib, Xiong, and Kira]{ma2019regretful}
C.-Y. Ma, Z.~Wu, G.~AlRegib, C.~Xiong, and Z.~Kira.
\newblock The regretful agent: Heuristic-aided navigation through progress
  estimation.
\newblock In \emph{Proceedings of the IEEE/CVF Conference on Computer Vision
  and Pattern Recognition}, pages 6732--6740, 2019.

\bibitem[Zadaianchuk et~al.(2022)Zadaianchuk, Martius, and
  Yang]{zadaianchuk2022self}
A.~Zadaianchuk, G.~Martius, and F.~Yang.
\newblock Self-supervised reinforcement learning with independently
  controllable subgoals.
\newblock In \emph{Conference on Robot Learning}. PMLR, 2022.

\bibitem[Suglia et~al.(2021)Suglia, Gao, Thomason, Thattai, and
  Sukhatme]{suglia2021embodied}
A.~Suglia, Q.~Gao, J.~Thomason, G.~Thattai, and G.~Sukhatme.
\newblock Embodied bert: A transformer model for embodied, language-guided
  visual task completion.
\newblock \emph{arXiv preprint arXiv:2108.04927}, 2021.

\bibitem[Nguyen et~al.(2021)Nguyen, Suganuma, and Okatani]{lwit}
V.-Q. Nguyen, M.~Suganuma, and T.~Okatani.
\newblock Look wide and interpret twice: Improving performance on interactive
  instruction-following tasks.
\newblock \emph{arXiv preprint arXiv:2106.00596}, 2021.

\bibitem[Kim et~al.(2021)Kim, Bhambri, Singh, Mottaghi, and Choi]{abp}
B.~Kim, S.~Bhambri, K.~P. Singh, R.~Mottaghi, and J.~Choi.
\newblock Agent with the big picture: Perceiving surroundings for interactive
  instruction following.
\newblock In \emph{Embodied AI Workshop CVPR}, 2021.

\bibitem[Song et~al.(2022)Song, Kil, Pan, Sadler, Chao, and Su]{song2022one}
C.~H. Song, J.~Kil, T.-Y. Pan, B.~M. Sadler, W.-L. Chao, and Y.~Su.
\newblock One step at a time: Long-horizon vision-and-language navigation with
  milestones.
\newblock \emph{arXiv preprint arXiv:2202.07028}, 2022.

\bibitem[Zhang and Chai(2021)]{hitut}
Y.~Zhang and J.~Chai.
\newblock Hierarchical task learning from language instructions with unified
  transformers and self-monitoring.
\newblock \emph{arXiv preprint arXiv:2106.03427}, 2021.

\bibitem[Chaplot et~al.(2020{\natexlab{a}})Chaplot, Gandhi, Gupta, and
  Salakhutdinov]{chaplot2020object}
D.~S. Chaplot, D.~P. Gandhi, A.~Gupta, and R.~R. Salakhutdinov.
\newblock Object goal navigation using goal-oriented semantic exploration.
\newblock \emph{Advances in Neural Information Processing Systems}, 33,
  2020{\natexlab{a}}.

\bibitem[Chaplot et~al.(2020{\natexlab{b}})Chaplot, Gandhi, Gupta, Gupta, and
  Salakhutdinov]{chaplot2020learning}
D.~S. Chaplot, D.~Gandhi, S.~Gupta, A.~Gupta, and R.~Salakhutdinov.
\newblock Learning to explore using active neural slam.
\newblock \emph{arXiv preprint arXiv:2004.05155}, 2020{\natexlab{b}}.

\bibitem[Li et~al.(2022)Li, Xia, Mart{\'\i}n-Mart{\'\i}n, Lingelbach,
  Srivastava, Shen, Vainio, Gokmen, Dharan, Jain, et~al.]{li2022igibson}
C.~Li, F.~Xia, R.~Mart{\'\i}n-Mart{\'\i}n, M.~Lingelbach, S.~Srivastava,
  B.~Shen, K.~E. Vainio, C.~Gokmen, G.~Dharan, T.~Jain, et~al.
\newblock igibson 2.0: Object-centric simulation for robot learning of everyday
  household tasks.
\newblock In \emph{Conference on Robot Learning}. PMLR, 2022.

\bibitem[Das et~al.(2018)Das, Datta, Gkioxari, Lee, Parikh, and
  Batra]{das2018embodied}
A.~Das, S.~Datta, G.~Gkioxari, S.~Lee, D.~Parikh, and D.~Batra.
\newblock Embodied question answering.
\newblock In \emph{Proceedings of the IEEE Conference on Computer Vision and
  Pattern Recognition}, pages 1--10, 2018.

\bibitem[Gordon et~al.(2018)Gordon, Kembhavi, Rastegari, Redmon, Fox, and
  Farhadi]{gordon2018iqa}
D.~Gordon, A.~Kembhavi, M.~Rastegari, J.~Redmon, D.~Fox, and A.~Farhadi.
\newblock Iqa: Visual question answering in interactive environments.
\newblock In \emph{Proceedings of the IEEE conference on computer vision and
  pattern recognition}, pages 4089--4098, 2018.

\bibitem[Liao et~al.(2019)Liao, Puig, Boben, Torralba, and
  Fidler]{liao2019synthesizing}
Y.-H. Liao, X.~Puig, M.~Boben, A.~Torralba, and S.~Fidler.
\newblock Synthesizing environment-aware activities via activity sketches.
\newblock In \emph{Proceedings of the IEEE/CVF Conference on Computer Vision
  and Pattern Recognition}, pages 6291--6299, 2019.

\bibitem[Trivedi et~al.(2021)Trivedi, Zhang, Sun, and Lim]{trivedi2021learning}
D.~Trivedi, J.~Zhang, S.-H. Sun, and J.~J. Lim.
\newblock Learning to synthesize programs as interpretable and generalizable
  policies.
\newblock \emph{Advances in neural information processing systems},
  34:\penalty0 25146--25163, 2021.

\bibitem[Wang et~al.(2020)Wang, Mao, Gershman, and Wu]{wang2020language}
R.~Wang, J.~Mao, S.~J. Gershman, and J.~Wu.
\newblock Language-mediated, object-centric representation learning.
\newblock \emph{arXiv preprint arXiv:2012.15814}, 2020.

\bibitem[Bisk et~al.(2020)Bisk, Holtzman, Thomason, Andreas, Bengio, Chai,
  Lapata, Lazaridou, May, Nisnevich, et~al.]{bisk2020experience}
Y.~Bisk, A.~Holtzman, J.~Thomason, J.~Andreas, Y.~Bengio, J.~Chai, M.~Lapata,
  A.~Lazaridou, J.~May, A.~Nisnevich, et~al.
\newblock Experience grounds language.
\newblock In \emph{Proceedings of the 2020 Conference on Empirical Methods in
  Natural Language Processing (EMNLP)}, 2020.

\bibitem[Prabhudesai et~al.(2020)Prabhudesai, Tung, Javed, Sieb, Harley, and
  Fragkiadaki]{prabhudesai2020embodied}
M.~Prabhudesai, H.-Y.~F. Tung, S.~A. Javed, M.~Sieb, A.~W. Harley, and
  K.~Fragkiadaki.
\newblock Embodied language grounding with 3d visual feature representations.
\newblock In \emph{Proceedings of the IEEE/CVF Conference on Computer Vision
  and Pattern Recognition}, pages 2220--2229, 2020.

\bibitem[He et~al.(2017)He, Gkioxari, Doll{\'a}r, and Girshick]{he2017mask}
K.~He, G.~Gkioxari, P.~Doll{\'a}r, and R.~Girshick.
\newblock Mask r-cnn.
\newblock In \emph{Proceedings of the IEEE international conference on computer
  vision}, pages 2961--2969, 2017.

\bibitem[Ding et~al.(2020)Ding, Huo, Yi, Wang, Shi, Lu, and
  Luo]{ding2020learning}
M.~Ding, Y.~Huo, H.~Yi, Z.~Wang, J.~Shi, Z.~Lu, and P.~Luo.
\newblock Learning depth-guided convolutions for monocular 3d object detection.
\newblock In \emph{CVPR}, pages 1000--1001, 2020.

\bibitem[Ding et~al.(2022)Ding, Xiao, Codella, Luo, Wang, and
  Yuan]{ding2022davit}
M.~Ding, B.~Xiao, N.~Codella, P.~Luo, J.~Wang, and L.~Yuan.
\newblock Davit: Dual attention vision transformers.
\newblock In \emph{ECCV}, 2022.

\bibitem[Kong et~al.(2014)Kong, Lin, Bansal, Urtasun, and Fidler]{kong2014you}
C.~Kong, D.~Lin, M.~Bansal, R.~Urtasun, and S.~Fidler.
\newblock What are you talking about? text-to-image coreference.
\newblock In \emph{CVPR}, 2014.

\bibitem[Plummer et~al.(2015)Plummer, Wang, Cervantes, Caicedo, Hockenmaier,
  and Lazebnik]{plummer2015flickr30k}
B.~A. Plummer, L.~Wang, C.~M. Cervantes, J.~C. Caicedo, J.~Hockenmaier, and
  S.~Lazebnik.
\newblock Flickr30k entities: Collecting region-to-phrase correspondences for
  richer image-to-sentence models.
\newblock In \emph{ICCV}, 2015.

\bibitem[Matuszek et~al.(2012)Matuszek, FitzGerald, Zettlemoyer, Bo, and
  Fox]{matuszek2012joint}
C.~Matuszek, N.~FitzGerald, L.~Zettlemoyer, L.~Bo, and D.~Fox.
\newblock A joint model of language and perception for grounded attribute
  learning.
\newblock In \emph{ICML}, 2012.

\bibitem[Karpathy and Fei-Fei(2015)]{karpathy2015deep}
A.~Karpathy and L.~Fei-Fei.
\newblock Deep visual-semantic alignments for generating image descriptions.
\newblock In \emph{CVPR}, 2015.

\bibitem[Mao et~al.(2016)Mao, Huang, Toshev, Camburu, Yuille, and
  Murphy]{mao2016generation}
J.~Mao, J.~Huang, A.~Toshev, O.~Camburu, A.~L. Yuille, and K.~Murphy.
\newblock Generation and comprehension of unambiguous object descriptions.
\newblock In \emph{CVPR}, 2016.

\bibitem[Zhang et~al.(2018)Zhang, Niu, and Chang]{zhang2018grounding}
H.~Zhang, Y.~Niu, and S.-F. Chang.
\newblock Grounding referring expressions in images by variational context.
\newblock In \emph{CVPR}, 2018.

\bibitem[Yang et~al.(2021)Yang, Tung, Zhang, Pathak, Pokle, Atkeson, and
  Fragkiadaki]{yang2021visually}
J.~Yang, H.-Y. Tung, Y.~Zhang, G.~Pathak, A.~Pokle, C.~G. Atkeson, and
  K.~Fragkiadaki.
\newblock Visually-grounded library of behaviors for manipulating diverse
  objects across diverse configurations and views.
\newblock In \emph{5th Annual Conference on Robot Learning}, 2021.

\bibitem[Chen et~al.(2020)Chen, Wang, Ma, Wong, and Wu]{chen2020cops}
Z.~Chen, P.~Wang, L.~Ma, K.-Y.~K. Wong, and Q.~Wu.
\newblock Cops-ref: A new dataset and task on compositional referring
  expression comprehension.
\newblock In \emph{Proceedings of the IEEE/CVF Conference on Computer Vision
  and Pattern Recognition}, pages 10086--10095, 2020.

\bibitem[Ding et~al.(2023)Ding, Shen, Fan, Chen, Chen, Luo, Tenenbaum, and
  Gan]{ding2023visual}
M.~Ding, Y.~Shen, L.~Fan, Z.~Chen, Z.~Chen, P.~Luo, J.~B. Tenenbaum, and
  C.~Gan.
\newblock Visual dependency transformers: Dependency tree emerges from reversed
  attention.
\newblock \emph{arXiv preprint arXiv:2304.03282}, 2023.

\bibitem[Mao et~al.(2019)Mao, Gan, Kohli, Tenenbaum, and
  Wu]{Mao2019NeuroSymbolic}
J.~Mao, C.~Gan, P.~Kohli, J.~B. Tenenbaum, and J.~Wu.
\newblock {The Neuro-Symbolic Concept Learner: Interpreting Scenes, Words, and
  Sentences From Natural Supervision}.
\newblock In \emph{ICLR}, 2019.

\bibitem[Chen et~al.(2021)Chen, Mao, Wu, Wong, Tenenbaum, and
  Gan]{zfchen2021iclr}
Z.~Chen, J.~Mao, J.~Wu, K.-Y.~K. Wong, J.~B. Tenenbaum, and C.~Gan.
\newblock Grounding physical concepts of objects and events through dynamic
  visual reasoning.
\newblock In \emph{ICLR}, 2021.

\bibitem[Mao et~al.(2021)Mao, Shi, Wu, Levy, and Tenenbaum]{mao2021grammar}
J.~Mao, F.~Shi, J.~Wu, R.~Levy, and J.~Tenenbaum.
\newblock Grammar-based grounded lexicon learning.
\newblock \emph{Advances in Neural Information Processing Systems}, 2021.

\bibitem[Bergen and Feldman(2008)]{bergen2008embodied}
B.~Bergen and J.~Feldman.
\newblock Embodied concept learning.
\newblock In \emph{Handbook of Cognitive Science}, pages 313--331. Elsevier,
  2008.

\bibitem[Hermann et~al.(2017)Hermann, Hill, Green, Wang, Faulkner, Soyer,
  Szepesvari, Czarnecki, Jaderberg, Teplyashin, et~al.]{hermann2017grounded}
K.~M. Hermann, F.~Hill, S.~Green, F.~Wang, R.~Faulkner, H.~Soyer,
  D.~Szepesvari, W.~M. Czarnecki, M.~Jaderberg, D.~Teplyashin, et~al.
\newblock Grounded language learning in a simulated 3d world.
\newblock \emph{arXiv}, 2017.

\bibitem[Chen et~al.(2022)Chen, Yi, Li, Ding, Torralba, Tenenbaum, and
  Gan]{chen2021comphy}
Z.~Chen, K.~Yi, Y.~Li, M.~Ding, A.~Torralba, J.~B. Tenenbaum, and C.~Gan.
\newblock Comphy: Compositional physical reasoning of objects and events from
  videos.
\newblock In \emph{International Conference on Learning Representations}, 2022.

\bibitem[Ding et~al.(2021)Ding, Chen, Du, Luo, Tenenbaum, and
  Gan]{ding2021dynamic}
M.~Ding, Z.~Chen, T.~Du, P.~Luo, J.~Tenenbaum, and C.~Gan.
\newblock Dynamic visual reasoning by learning differentiable physics models
  from video and language.
\newblock \emph{Advances in Neural Information Processing Systems}, 34, 2021.

\bibitem[Feng et~al.(2021)Feng, Li, Li, Zhang, Zhang, Zhu, Zhang, Wang, and
  Mian]{feng2021free}
M.~Feng, Z.~Li, Q.~Li, L.~Zhang, X.~Zhang, G.~Zhu, H.~Zhang, Y.~Wang, and
  A.~Mian.
\newblock Free-form description guided 3d visual graph network for object
  grounding in point cloud.
\newblock In \emph{ICCV}, 2021.

\bibitem[Roh et~al.(2022)Roh, Desingh, Farhadi, and Fox]{roh2022languagerefer}
J.~Roh, K.~Desingh, A.~Farhadi, and D.~Fox.
\newblock Languagerefer: Spatial-language model for 3d visual grounding.
\newblock In \emph{Conference on Robot Learning}. PMLR, 2022.

\bibitem[Achlioptas et~al.(2020)Achlioptas, Abdelreheem, Xia, Elhoseiny, and
  Guibas]{achlioptas2020referit3d}
P.~Achlioptas, A.~Abdelreheem, F.~Xia, M.~Elhoseiny, and L.~Guibas.
\newblock Referit3d: Neural listeners for fine-grained 3d object identification
  in real-world scenes.
\newblock In \emph{ECCV}, 2020.

\bibitem[Lewis et~al.(2020)Lewis, Liu, Goyal, Ghazvininejad, Mohamed, Levy,
  Stoyanov, and Zettlemoyer]{lewis-etal-2020-bart}
M.~Lewis, Y.~Liu, N.~Goyal, M.~Ghazvininejad, A.~Mohamed, O.~Levy, V.~Stoyanov,
  and L.~Zettlemoyer.
\newblock {BART}: Denoising sequence-to-sequence pre-training for natural
  language generation, translation, and comprehension.
\newblock In \emph{Proceedings of the 58th Annual Meeting of the Association
  for Computational Linguistics}, pages 7871--7880, Online, July 2020.
  Association for Computational Linguistics.
\newblock \doi{10.18653/v1/2020.acl-main.703}.
\newblock URL \url{https://aclanthology.org/2020.acl-main.703}.

\bibitem[Meelfy(2019)]{meelfy}
Meelfy.
\newblock Pytorch\_pretrained\_bert, 2019.
\newblock URL \url{https://github.com/Meelfy/pytorch_pretrained_BERT}.

\bibitem[Carbon(2014)]{carbon2014understanding}
C.-C. Carbon.
\newblock Understanding human perception by human-made illusions.
\newblock \emph{Frontiers in human neuroscience}, 8:\penalty0 566, 2014.

\bibitem[Regan(2000)]{regan2000human}
D.~Regan.
\newblock \emph{Human perception of objects}.
\newblock Sunderland, MA: Sinauer, 2000.

\bibitem[Ba et~al.(2016)Ba, Kiros, and Hinton]{ba2016layer}
J.~L. Ba, J.~R. Kiros, and G.~E. Hinton.
\newblock Layer normalization.
\newblock \emph{arXiv preprint arXiv:1607.06450}, 2016.

\bibitem[Hendrycks and Gimpel(2016)]{hendrycks2016gaussian}
D.~Hendrycks and K.~Gimpel.
\newblock Gaussian error linear units (gelus).
\newblock \emph{arXiv preprint arXiv:1606.08415}, 2016.

\bibitem[Kuhn(1955)]{kuhn1955hungarian}
H.~W. Kuhn.
\newblock The hungarian method for the assignment problem.
\newblock \emph{Naval research logistics quarterly}, 2\penalty0 (1-2):\penalty0
  83--97, 1955.

\bibitem[Maguire et~al.(1998)Maguire, Burgess, Donnett, Frackowiak, Frith, and
  O'Keefe]{maguire1998knowing}
E.~A. Maguire, N.~Burgess, J.~G. Donnett, R.~S. Frackowiak, C.~D. Frith, and
  J.~O'Keefe.
\newblock Knowing where and getting there: a human navigation network.
\newblock \emph{Science}, 280\penalty0 (5365):\penalty0 921--924, 1998.

\bibitem[Godard et~al.(2019)Godard, Mac~Aodha, Firman, and
  Brostow]{godard2019digging}
C.~Godard, O.~Mac~Aodha, M.~Firman, and G.~J. Brostow.
\newblock Digging into self-supervised monocular depth estimation.
\newblock In \emph{Proceedings of the IEEE/CVF International Conference on
  Computer Vision}, pages 3828--3838, 2019.

\bibitem[Zhou et~al.(2017)Zhou, Brown, Snavely, and Lowe]{zhou2017unsupervised}
T.~Zhou, M.~Brown, N.~Snavely, and D.~G. Lowe.
\newblock Unsupervised learning of depth and ego-motion from video.
\newblock In \emph{Proceedings of the IEEE conference on computer vision and
  pattern recognition}, pages 1851--1858, 2017.

\bibitem[Kendall and Gal(2017)]{kendall2017uncertainties}
A.~Kendall and Y.~Gal.
\newblock What uncertainties do we need in bayesian deep learning for computer
  vision?
\newblock \emph{Advances in neural information processing systems}, 30, 2017.

\bibitem[Sethian(1996)]{Sethian1591}
J.~A. Sethian.
\newblock A fast marching level set method for monotonically advancing fronts.
\newblock \emph{Proceedings of the National Academy of Sciences}, 93\penalty0
  (4):\penalty0 1591--1595, 1996.
\newblock ISSN 0027-8424.
\newblock \doi{10.1073/pnas.93.4.1591}.
\newblock URL \url{https://www.pnas.org/content/93/4/1591}.

\bibitem[Tran et~al.(2015)Tran, Bourdev, Fergus, Torresani, and
  Paluri]{tran2015learning}
D.~Tran, L.~Bourdev, R.~Fergus, L.~Torresani, and M.~Paluri.
\newblock Learning spatiotemporal features with 3d convolutional networks.
\newblock In \emph{Proceedings of the IEEE international conference on computer
  vision}, pages 4489--4497, 2015.

\bibitem[Kolve et~al.(2019)Kolve, Mottaghi, Han, VanderBilt, Weihs, Herrasti,
  Gordon, Zhu, Gupta, and Farhadi]{ai2thor}
E.~Kolve, R.~Mottaghi, W.~Han, E.~VanderBilt, L.~Weihs, A.~Herrasti, D.~Gordon,
  Y.~Zhu, A.~Gupta, and A.~Farhadi.
\newblock Ai2-thor: An interactive 3d environment for visual ai, 2019.

\bibitem[Hui(2022)]{hui2022rm}
T.-W. Hui.
\newblock Rm-depth: Unsupervised learning of recurrent monocular depth in
  dynamic scenes.
\newblock In \emph{Proceedings of the IEEE/CVF Conference on Computer Vision
  and Pattern Recognition}, pages 1675--1684, 2022.

\bibitem[Luo et~al.(2020)Luo, Huang, Szeliski, Matzen, and
  Kopf]{luo2020consistent}
X.~Luo, J.-B. Huang, R.~Szeliski, K.~Matzen, and J.~Kopf.
\newblock Consistent video depth estimation.
\newblock \emph{ACM Transactions on Graphics (ToG)}, 39\penalty0 (4):\penalty0
  71--1, 2020.

\bibitem[Ranjan et~al.(2019)Ranjan, Jampani, Balles, Kim, Sun, Wulff, and
  Black]{ranjan2019competitive}
A.~Ranjan, V.~Jampani, L.~Balles, K.~Kim, D.~Sun, J.~Wulff, and M.~J. Black.
\newblock Competitive collaboration: Joint unsupervised learning of depth,
  camera motion, optical flow and motion segmentation.
\newblock In \emph{Proceedings of the IEEE/CVF conference on computer vision
  and pattern recognition}, pages 12240--12249, 2019.

\bibitem[Gordon et~al.(2019)Gordon, Li, Jonschkowski, and
  Angelova]{gordon2019depth}
A.~Gordon, H.~Li, R.~Jonschkowski, and A.~Angelova.
\newblock Depth from videos in the wild: Unsupervised monocular depth learning
  from unknown cameras.
\newblock In \emph{Proceedings of the IEEE/CVF International Conference on
  Computer Vision}, pages 8977--8986, 2019.

\bibitem[Yao et~al.(2018)Yao, Luo, Li, Fang, and Quan]{yao2018mvsnet}
Y.~Yao, Z.~Luo, S.~Li, T.~Fang, and L.~Quan.
\newblock Mvsnet: Depth inference for unstructured multi-view stereo.
\newblock In \emph{Proceedings of the European Conference on Computer Vision
  (ECCV)}, pages 767--783, 2018.

\bibitem[Lu et~al.(2021)Lu, Xu, Ding, Lu, and Xiang]{lu2021global}
Y.~Lu, X.~Xu, M.~Ding, Z.~Lu, and T.~Xiang.
\newblock A global occlusion-aware approach to self-supervised monocular visual
  odometry.
\newblock In \emph{Proceedings of the AAAI Conference on Artificial
  Intelligence}, volume~35, pages 2260--2268, 2021.

\bibitem[Guo et~al.(2019)Guo, Yang, Yang, Wang, and Li]{guo2019group}
X.~Guo, K.~Yang, W.~Yang, X.~Wang, and H.~Li.
\newblock Group-wise correlation stereo network.
\newblock In \emph{Proceedings of the IEEE/CVF Conference on Computer Vision
  and Pattern Recognition}, pages 3273--3282, 2019.

\bibitem[Wang et~al.(2021)Wang, Galliani, Vogel, Speciale, and
  Pollefeys]{wang2021patchmatchnet}
F.~Wang, S.~Galliani, C.~Vogel, P.~Speciale, and M.~Pollefeys.
\newblock Patchmatchnet: Learned multi-view patchmatch stereo.
\newblock In \emph{Proceedings of the IEEE/CVF Conference on Computer Vision
  and Pattern Recognition}, pages 14194--14203, 2021.

\bibitem[Tang et~al.(2020)Tang, Tian, Feng, Li, and Tan]{tang2020learning}
J.~Tang, F.-P. Tian, W.~Feng, J.~Li, and P.~Tan.
\newblock Learning guided convolutional network for depth completion.
\newblock \emph{IEEE Transactions on Image Processing}, 30:\penalty0
  1116--1129, 2020.

\bibitem[Xu et~al.(2019)Xu, Zhu, Shi, Zhang, Bao, and Li]{xu2019depth}
Y.~Xu, X.~Zhu, J.~Shi, G.~Zhang, H.~Bao, and H.~Li.
\newblock Depth completion from sparse lidar data with depth-normal
  constraints.
\newblock In \emph{Proceedings of the IEEE/CVF International Conference on
  Computer Vision}, pages 2811--2820, 2019.

\bibitem[Laina et~al.(2016)Laina, Rupprecht, Belagiannis, Tombari, and
  Navab]{laina2016deeper}
I.~Laina, C.~Rupprecht, V.~Belagiannis, F.~Tombari, and N.~Navab.
\newblock Deeper depth prediction with fully convolutional residual networks.
\newblock In \emph{2016 Fourth international conference on 3D vision (3DV)},
  pages 239--248. IEEE, 2016.

\bibitem[Cao et~al.(2017)Cao, Wu, and Shen]{cao2017estimating}
Y.~Cao, Z.~Wu, and C.~Shen.
\newblock Estimating depth from monocular images as classification using deep
  fully convolutional residual networks.
\newblock \emph{IEEE Transactions on Circuits and Systems for Video
  Technology}, 28\penalty0 (11):\penalty0 3174--3182, 2017.

\bibitem[Yang et~al.(2020)Yang, Stumberg, Wang, and Cremers]{yang2020d3vo}
N.~Yang, L.~v. Stumberg, R.~Wang, and D.~Cremers.
\newblock D3vo: Deep depth, deep pose and deep uncertainty for monocular visual
  odometry.
\newblock In \emph{Proceedings of the IEEE/CVF Conference on Computer Vision
  and Pattern Recognition}, pages 1281--1292, 2020.

\bibitem[Li et~al.(2021)Li, Huang, Liu, Zou, and Yu]{li2021structdepth}
B.~Li, Y.~Huang, Z.~Liu, D.~Zou, and W.~Yu.
\newblock Structdepth: Leveraging the structural regularities for
  self-supervised indoor depth estimation.
\newblock In \emph{Proceedings of the IEEE/CVF International Conference on
  Computer Vision}, pages 12663--12673, 2021.

\bibitem[Ji et~al.(2021)Ji, Li, Bhanu, and Xu]{ji2021monoindoor}
P.~Ji, R.~Li, B.~Bhanu, and Y.~Xu.
\newblock Monoindoor: Towards good practice of self-supervised monocular depth
  estimation for indoor environments.
\newblock In \emph{Proceedings of the IEEE/CVF International Conference on
  Computer Vision}, pages 12787--12796, 2021.

\bibitem[Pillai et~al.(2019)Pillai, Ambru{\c{s}}, and
  Gaidon]{pillai2019superdepth}
S.~Pillai, R.~Ambru{\c{s}}, and A.~Gaidon.
\newblock Superdepth: Self-supervised, super-resolved monocular depth
  estimation.
\newblock In \emph{2019 International Conference on Robotics and Automation
  (ICRA)}, pages 9250--9256. IEEE, 2019.

\bibitem[Mur-Artal et~al.(2015)Mur-Artal, Montiel, and Tardos]{mur2015orb}
R.~Mur-Artal, J.~M.~M. Montiel, and J.~D. Tardos.
\newblock Orb-slam: a versatile and accurate monocular slam system.
\newblock \emph{IEEE transactions on robotics}, 31\penalty0 (5):\penalty0
  1147--1163, 2015.

\bibitem[Izadi et~al.(2011)Izadi, Kim, Hilliges, Molyneaux, Newcombe, Kohli,
  Shotton, Hodges, Freeman, Davison, et~al.]{izadi2011kinectfusion}
S.~Izadi, D.~Kim, O.~Hilliges, D.~Molyneaux, R.~Newcombe, P.~Kohli, J.~Shotton,
  S.~Hodges, D.~Freeman, A.~Davison, et~al.
\newblock Kinectfusion: real-time 3d reconstruction and interaction using a
  moving depth camera.
\newblock In \emph{Proceedings of the 24th annual ACM symposium on User
  interface software and technology}, pages 559--568, 2011.

\bibitem[Shan and Englot(2018)]{shan2018lego}
T.~Shan and B.~Englot.
\newblock Lego-loam: Lightweight and ground-optimized lidar odometry and
  mapping on variable terrain.
\newblock In \emph{2018 IEEE/RSJ International Conference on Intelligent Robots
  and Systems (IROS)}, pages 4758--4765. IEEE, 2018.

\bibitem[Klein and Murray(2007)]{klein2007parallel}
G.~Klein and D.~Murray.
\newblock Parallel tracking and mapping for small ar workspaces.
\newblock In \emph{2007 6th IEEE and ACM international symposium on mixed and
  augmented reality}, pages 225--234. IEEE, 2007.

\bibitem[Zhang and Singh(2014)]{zhang2014loam}
J.~Zhang and S.~Singh.
\newblock Loam: Lidar odometry and mapping in real-time.
\newblock In \emph{Robotics: Science and Systems}, volume~2, pages 1--9.
  Berkeley, CA, 2014.

\end{thebibliography}
